\documentclass[11pt]{article}

\usepackage{fullpage}
\usepackage{amsmath}
\usepackage{amsthm,amsfonts,amssymb}
\usepackage{xspace}
\usepackage{bbm}
\usepackage{subcaption}
\usepackage{graphicx}
\usepackage{bbm}
\usepackage{hyperref}   
\usepackage{etoolbox}
\usepackage{enumitem}
\usepackage{authblk}

\usepackage[noend]{algorithmic}
\usepackage[ruled,vlined]{algorithm2e}
\usepackage{comment}
\usepackage{natbib}
\usepackage{float}

	\hypersetup{
			colorlinks=true,
			linkcolor=[rgb]{0,0,.5},
			urlcolor=[rgb]{0,0,.5},
			citecolor=[rgb]{0,0,.5},
			pdfstartview=FitH}

\usepackage{accents}

\usepackage{graphicx} 
\usepackage{booktabs} 

\usepackage{caption}

\usepackage{xcolor}

\usepackage{pgfplots}

\pgfplotsset{compat=1.10}

\usepackage{tikz}

\usepackage{enumitem}
\usepackage{mathtools}
\usetikzlibrary{quotes,positioning}
\allowdisplaybreaks

\newtheorem{lemma}{Lemma}[section]

\newtheorem{definition}{Definition}[section]
\newtheorem{remark}{Remark}[section]
\newtheorem{example}{Example}[section]

\newtheorem{obs}{Observation}[section]

\newcounter{numrellocal}
\renewcommand{\thenumrellocal}{\arabic{numrellocal}}
\newcounter{numrelglobal}

\makeatletter
\newcommand{\numrel}[2]{
  \stepcounter{numrellocal}
  \refstepcounter{numrelglobal}
  \ltx@label{#2}
  \overset{(\thenumrellocal)}{#1}
}
\makeatother

\usepackage{xcolor}
\newcommand{\cj}[1]{\textcolor{blue}{[CJ: #1]}}

\newcommand{\ind}{\mathbbm{1}}
\newcommand{\R}{\mathbb{R}}
\newcommand{\E}{\mathop{\mathbb{E}}}

\newcommand{\cX}{\mathcal{X}}
\newcommand{\cY}{\mathcal{Y}}

\newcommand{\cP}{\mathcal{P}}

\newcommand{\cG}{\mathcal{G}}

\newcommand{\cT}{\mathcal{T}}
\newcommand{\Tau}{\mathcal{T}}

\newcommand{\bN}{\mathbb{N}}

\newcommand{\cD}{\mathcal{D}}

\newcommand{\bH}{\overline{H}}

\newcommand{\Binv}{B^{-1}}
\newcommand{\Bm}{B_m}
\newcommand{\Bminv}{B_m^{-1}}

\newcommand{\cover}{\mathrm{Cover}}

\usepackage{thmtools} 
\usepackage{thm-restate}

\usepackage{amsmath}
\usepackage{pgfplots}

\title{Practical Adversarial Multivalid Conformal Prediction
}

\author[1]{Osbert Bastani}
\author[1]{Varun Gupta}
\author[1]{Christopher Jung}
\author[1]{Georgy Noarov}
\author[1]{Ramya Ramalingam}
\author[1]{Aaron Roth}
\affil[1]{Department of Computer and Information Sciences, University of Pennsylvania}

\begin{document}
\maketitle

\begin{abstract}
We give a simple, generic conformal prediction method  for sequential prediction that  achieves target empirical coverage guarantees against adversarially chosen data. It is computationally lightweight --- comparable to split conformal prediction --- but does not require having a held-out validation set, and so all data can be used for training models from which to derive a conformal score. It gives stronger than marginal coverage guarantees in two ways. First, it gives \emph{threshold calibrated} prediction sets that have correct empirical coverage even conditional on the threshold used to form the prediction set from the conformal score. Second, the user can specify an arbitrary collection of subsets of the feature space --- possibly intersecting --- and the coverage guarantees also hold conditional on membership in each of these subsets. We call our algorithm MVP, short for MultiValid Prediction. We give both theory and an extensive set of empirical evaluations. 
\end{abstract}

\section{Introduction}
Consider the problem of predicting labels $y\in\mathcal{Y}$ given examples $x\in\mathcal{X}$. One popular strategy for expressing uncertainty is to allow the algorithm to produce a \emph{prediction set} $\Tau \subseteq\mathcal{Y}$ rather than an individual label. We give a simple, practical algorithm for producing prediction sets in sequential prediction problems over an arbitrary domain $\cX\times\cY$, given any data-dependent sequence of conformal score functions $s_t:\cX \times \cY \rightarrow \mathbb{R}_{\geq 0}$. In rounds $t$, an example represented by a feature vector $x_t \in \cX$ arrives. We can define an  arbitrary conformal score  $s_t:\cX \times \cY \rightarrow \mathbb{R}_{\geq 0}$ that can  depend on previously observed examples in arbitrary ways. We produce a round-dependent threshold $q_t$, which gives us a prediction set $\Tau_t = \{y \in \cY : s_t(x_t,y) \leq q_t\}$. We then learn the true label $y_t$, and we say our prediction set \emph{covers} $y_t$ if $y_t \in \Tau_t$.  Given a coverage target $1-\delta$, our goal is to produce intervals that have correct  empirical coverage --- i.e. that cover a $1-\delta$ fraction of the labels (we do not want either over-coverage or under-coverage). We wish to make as few assumptions as possible, so that our method is \emph{robust} to arbitrary and unanticipated distribution shift, and applies to e.g. time series data which are very far from exchangeable. We also want our coverage guarantees to be meaningful not just marginally, but at finer granularities: \emph{conditional} on both the threshold value we choose, and on membership of $x_t \in G$ for a set of groups $G \in \cG$ that can be arbitrarily defined and intersecting. Finally, we want our algorithm to have low computational overhead, so that it can be applied as a wrapper on top of arbitrary prediction methods, for both regression and classification. The algorithm we give achieves these goals and has a number of desirable properties:

\paragraph{Worst-Case Empirical Coverage:} 
Our method has worst case \emph{adversarial} guarantees. The sequence of examples $\{(x_t,y_t)\}_{t=1}^T$ does not need to be drawn from an exchangeable distribution as it does for standard conformal prediction methods \citep{conformal} --- instead it can be  chosen by an adaptive adversary.  The conformal scores $s_t$ can be arbitrary and can depend on data from previous rounds (e.g. they can be derived from models that have been trained on all past data, and so there is no need to separate data into a training and calibration set as in split conformal prediction \citep{lei2018distribution}). Thus our method can tolerate time series data as well as arbitrary and unanticipated distribution shift of any sort.

\paragraph{Calibrated, Multivalid Coverage} 
Our prediction sets obtain their target empirical coverage level not just marginally, but in a \emph{threshold-calibrated} fashion. This means that for every threshold $q$, the subsequence of rounds $t$ on which the threshold $q_t = q$ approaches the target empirical coverage\footnote{\label{footnote:allno}Calibration is especially important in a distribution free setting, when coverage is measured empirically. If we only asked for the target marginal empirical coverage as \cite{gibbs2021adaptive} do, rather than for threshold-calibrated prediction sets, it would be possible to obtain the right coverage rate by ``cheating'' in the following uninformative way: at each round, predict $S_t = \cY$ the full label set on a $1-\delta$ fraction of rounds (which is guaranteed to cover the label), and the empty set $S_t = \emptyset$ on the remaining $\delta$ fraction of rounds (which is guaranteed not to cover it). This obtains empirical coverage rate $1-\delta \pm O(1/T)$ marginally, but not conditional on the prediction sets chosen.}. We also promise group conditional coverage: we can specify an arbitrary collection of \emph{groups} $\cG$. Each group $G \in \cG$ represents an arbitrary subset of the feature space: $G \subseteq \cX$. These groups can intersect in arbitrary ways. For example, $\cG$ could represent collections of demographic groups based on race, age, income, or medical history, datapoints could represent people who are members of an arbitrary subset of these groups. Our method promises that simultaneously for each of these groups $G$, on the subsequence of rounds $t$ for which $x_t \in G$, our intervals obtain their target empirical coverage rate (again, in a calibrated fashion). 

\paragraph{Computationally Lightweight:} Our method is computationally lightweight: at each round, it only needs to enumerate candidate thresholds $q_t$ in some discrete range, the groups $G$ such that $x_t \in G$ and maintain records of the empirical coverage rate for each threshold and group. Hence it is comparable in cost to split conformal prediction methods \citep{lei2018distribution} despite its ability to use all data for model training. We give an implementation of our algorithm and an extensive empirical evaluation. In comparison, prior work which obtains comparable theoretical guarantees \citep{multivalid} (for the special case of prediction intervals in regression problems) does not give a practical algorithm --- the algorithm of \cite{multivalid} requires solving an exponentially large linear program at each round, using the Ellipsoid algorithm paired with a separation oracle.


\paragraph{Nearly Statistically Optimal Rates:} Threshold calibrated multivalid prediction sets require that simultaneously for each threshold $q$ and each group $G \in \cG$, the empirical coverage on the sequence of days for which $q_t = q$ and $x_t \in G$ approach the coverage target $1-\delta$. For each threshold $q$ and group $G$, let $n^{G,q}$ denote the length of this sequence. If we were in a setting where the labels $y_t$ were drawn from a known distribution, and our prediction sets had coverage probability exactly $1-\delta$ on the underlying distribution, we would still expect that our \emph{empirical} coverage on a subsequence defined by $q$ and $G$ would deviate from the target $1-\delta$ by a $\pm 1/\sqrt{n^{G,q}}$ term. The prior theoretical bound given by \cite{multivalid} has coverage guarantees that for each $q$ and $G$ differ from their target by $\tilde O(\sqrt{T}/n^{G,q})$, which is substantially sub-optimal for sequences such that $n^{G,q} \ll T$. Our algorithm promises coverage rates for each pair $(G,q)$ that differ from their target by an optimal $\tilde O(1/\sqrt{n^{G,q}})$ term. 

\paragraph{} We give an extensive experimental evaluation\footnote{The code to replicate all of our experiments can be found at \href{https://github.com/ProgBelarus/MultiValidPrediction }{https://github.com/ProgBelarus/MultiValidPrediction}} of our algorithm in a number of settings, and compare to  split conformal prediction \citep{lei2018distribution}, as well as prior work that is designed to handle limited forms of known distribution shift \citep{tibshirani2019conformal}, conservative forms of groupwise coverage \citep{barber2019limits}, and give adversarial (but uncalibrated) coverage guarantees \citep{gibbs2021adaptive}. In each setting, we show that our algorithm is competitive with previous work ``on their turf'' (i.e. in settings for which their assumptions are satisfied and we use their evaluation metrics). We then go on to show that our method gives substantial improvements when either the setting or the evaluation metric becomes more difficult --- e.g. when the distribution shift is unanticipated, when we measure group-wise rather than just marginal coverage, or when the data comes in adversarial ordering. In some cases we improve on standard techniques even in standard ``benign'' settings: for example, we improve on split conformal prediction in an online linear regression setting with i.i.d. data when the evaluation metric is just marginal coverage, but the regression function has to be learned from the same stream of data used to calibrate the prediction intervals. This is because split conformal prediction requires using separate splits of the data for training the regression and calibrating the prediction intervals to maintain exchangability of the conformal scores --- but since our method does not require exchangability, we are able to use all of the data for both tasks. 

\subsection{Additional Related Work}
 See \cite{angelopoulos2021gentle} for an excellent recent survey of conformal prediction. The weaknesses of these methods that we seek to address --- namely, that in the worst case they provide only marginal coverage, and that they rely on strong distributional assumptions (typically \emph{exchangeability}) --- have been noted before. For example, \cite{romano2020malice} note that marginal coverage guarantees are undesirable when making predictions about people, and give group conditional guarantees  for \emph{disjoint} groups by calibrating separately on each group. This fails when the groups can intersect. \cite{barber2019limits}  provide guarantees that are valid conditional on membership in intersecting subgroups $\cG$. They take a conservative approach, by computing prediction sets separately for each group, and then taking the union of all of these prediction sets over the  groups of a new individual. The result is that their prediction sets, unlike ours, are conservative and so do not approach their target coverage level, even in the limit. These results both require exchangeable data.   \cite{chernozhukov2018exact} consider the problem of conformal prediction for time series data, for which an exchangeability assumption may not hold. They show that if the data comes from a rapidly mixing process then it is still possible to obtain approximate marginal coverage guarantees.  \cite{tibshirani2019conformal} consider the problem of conformal prediction under \emph{covariate shift}, in which the marginal distribution on features $\cX$ differs between the training and test distributions, but the conditional distribution on labels $\cY | \cX$ remains the same. They show how to adapt techniques from conformal prediction when the changepoint is known, and the likelihood ratio between the training and test distribution is known. \cite{gibbs2021adaptive} give a method (ACI, for Adaptive Conformal Inference)  that can guarantee target marginal coverage without any assumptions on the data generating process.  \cite{aci2} and \cite{aci3} give refinements of the ACI procedure --- for example, \cite{aci3} dispenses with the need for a holdout set. In contrast to these papers, our prediction sets promise not just marginal coverage, but are  ``threshold-calibrated'' and hold also conditional on membership in arbitrary  sub-groups. \cite{vovk2002line} notes the importance of calibration and shows that conformal prediction methods are threshold calibrated in a strong sense when run on exchangeable data distributions.

Following a recent resurgence in interest in conformal techniques, a number of papers have proposed conformal scores that have desirable properties. For example, \cite{hoff2021bayes} gives a conformal score that produces prediction sets with Bayes optimal risk, given that an underlying Bayesian model is correctly specified. \cite{romano2019conformalized} gave a conformal score based on quantile regression that allows conformal prediction intervals in regression problems to be adaptive to heteroscedasticity. \cite{angelopoulos2020uncertainty} and \cite{romano2020classification} give methods in the classification setting for producing prediction sets whose size adapts to the (difficulty of the) example. Our work is complementary to this line of work: just like traditional methods of conformal prediction, we too take as input arbitrary conformal scores. Thus we can adopt any of these conformal score functions and inherit their properties, while providing the stronger worst-case guarantees of multivalid coverage. 

For the special case of prediction intervals, the type of \emph{multi-valid} prediction  that we study was first defined in \cite{momentmulti}, who gave a way of obtaining it in the \emph{batch} setting for i.i.d.\ data, via producing \emph{multicalibrated} estimates of label variances and higher moments.  \cite{multivalid} proved that there exists an online prediction algorithm that gives the sort of multi-valid prediction intervals that we consider in this work. The algorithm we give in this paper is both much more efficient (their algorithm was only of theoretical interest and involved solving an exponentially large linear program) and has substantially better (optimal) convergence bounds. 


Finally, the notion of multivalidity is related to subgroup fairness notions \citep{gerrymandering,subgroup,multicalibration,multiaccuracy} that ask for statistical ``fairness'' constraints of various sorts to hold across all subgroups defined by some rich class $\cG$. In particular, it is closely related to multicalibration \citep{multicalibration} which asks for point predictions that are calibrated not just marginally, but also conditionally on membership in a large number of intersecting demographic groups $\cG$.

\section{Preliminaries}
\subsection{Notation}
We let $\cX$ denote a feature domain and $\cY$ a label domain. We write $\cG \subseteq 2^\cX$ to denote a collection of subsets of $\cX$. Given any $x \in \cX$, we write $\cG(x)$ for the set of groups that contain $x$, i.e.\ $\cG(x) = \{G \in \cG: x \in G\}$. For any positive integer $T$, we write $[T] = \{1,\ldots,T\}$. In general, we denote random variables with tildes (e.g.\ $\tilde X$, $\tilde Y$) to distinguish them from their realizations (denoted e.g.\ $X$, $Y$). Given a set $A$, we write $\Delta A$ for the probability distribution over the elements in $A$. 

\subsection{Online Uncertainty Quantification}
Our uncertainty quantification is based on a bounded conformal score function $s_t: \cX \times \cY \to \R$ which can change in arbitrary ways between rounds $t \in [T]$. Without loss of generality\footnote{When $s_t(x,y) \in [L,U]$, then we can have the learner use the conformal score $s'_t(x,y) = \frac{s_t(x,y) - L}{U-L}$. When the learner produces a conformity threshold $q'_t$ with respect to the new scoring function $s'_t$, we may obtain a threshold $q_t$ with respect to the original score $s_t$ by setting $q_t = q'_t(U-L) + L$.}, we assume that the scoring function takes values in the unit interval: 
$s_t(x, y) \in [0,1]$ for any $x \in \cX, y \in \cY$, and $t\in [T]$. Fix some target coverage rate $1-\delta$. In each round $t \in [T]$, an interaction between a \emph{learner} and an \emph{adversary} proceeds as follows:
\begin{enumerate}
    \item The \emph{learner} chooses a conformal score function $s_t:\cX \times \cY \rightarrow [0, 1]$, which may be observed by the adversary.
    \item The \emph{adversary} chooses a joint distribution over feature vectors $x_t \in \cX$ and labels $y_t \in \cY$. The learner receives $x_t$ (a realized feature vector), but no information about the label $y_t$. 
    \item The learner produces a conformity threshold $q_t$. This corresponds to a prediction set which the learner outputs: \[
        \Tau_t(x_t) = \{y  \in \cY: s_t(x_t, y) \le q_t\}.
    \]
    \item The learner then learns the realized label $y_t$. 
\end{enumerate}

\begin{example}
There are many natural ways to form conformal score functions --- see  \citep{angelopoulos2021gentle} for an informative survey. Suppose we are in a regression setting and wish to quantify the uncertainty of a regression model $f_t:\cX\rightarrow [0,1]$, where $f_t$ is trained on all data points $(x_\tau,y_\tau)$ for $\tau < t$. We could then choose: $s_t(x,y) = |f_t(x) - y|$. With this choice of score function, the prediction set $\Tau_t(x_t)$ corresponds to a prediction interval centered at $f_t(x_t)$ of width $2q_t$. Alternately, for more informative intervals, we could  train a quantile regression model  as in \cite{romano2019conformalized} to obtain a regression function $f_t(x_t,\alpha)$ which attempts to estimate the $\alpha$-quantile of the label distribution conditional on $x$. We can then set $s_t(x,y) = \max(f_t(x_t,\delta/2)-y, y-f_t(x_t,1-\delta/2))$. In this case, prediction sets $\Tau_t(x_t)$ start from the guesses that $f_t$ makes as to a ``correct'' $1-\delta$ coverage interval, with width adjusted by $q_t$. Alternatively, suppose we are in a classification setting and have trained a model that given $x$ produces scores $f_t(x,y)$ for each $y \in \cY$ --- for example, the output of a softmax layer in a neural network. Following  \cite{angelopoulos2020uncertainty} and \cite{romano2020classification}, we could define a conformal score function $s_t(x,y) = \sum_{i=1}^{k(x,y)} f_t(x,\pi_i(x))$ where $\pi_i(x)$ is the label $y \in \cY$ that comes in $i$'th place when sorted by $f_t(x,y)$, and $k(x,y)$ is the index of label $y$ in this ordering. In this case, the prediction sets $\Tau_t(x_t)$ correspond to the prefix of labels sorted in order by their probability as estimated by $f_t$, where the length of the prefix is chosen so that their cumulative estimated probability is at most $q_t$. There are many other examples, and we can take advantage of any of them. 
\end{example}

Ideally, the learner wants to produce prediction sets $\Tau_t(x_t)$ that cover the true label $y$ with probability $1-\delta$ over the randomness of the adversary's unknown label distribution:
    $\Pr_{y | x_t}[y \in \Tau_t(x_t)] \approx 1-\delta$. Because of the structure of the prediction sets, this is equivalent to choosing a conformity threshold $q_t$ such that over the randomness of the adversary's unknown label distribution: $\Pr_{y | x_t}[s_t(x_t, y) \le q_t ] \approx 1-\delta$.

Because the adversary may choose the label distribution with knowledge of the conformal score function, we will elide the particulars of the conformal score function and the distribution on labels $y_t$ in our derivation, and instead equivalently imagine the adversary  directly choosing a distribution over conformal scores $s_t$ conditional on $x_t$ (representing the distribution over conformal scores $s_t(x_t,y_t)$). We may thus view the interaction in the following simplified form:
\begin{enumerate}
    \item The \emph{adversary} chooses a joint distribution over feature vector $x_t \in \cX$ and conformal score $s_t \in  [0, 1]$. The learner receives $x_t$ (a realized feature vector), but no information about $s_t$. 
    \item The learner produces a conformity threshold $q_t$.
    \item The learner observes the realized conformal score $s_t$. 
\end{enumerate}

For any round $t \in [T]$, we write $\pi_t = (x_t, s_t, q_t)$ to denote the realized outcomes in round $t$ and similarly write $\pi_{t':t}$ for the \emph{transcript} of the interaction between rounds $t' \leq \tau \leq t$: $\pi_{t':t} = ((x_\tau, s_\tau, q_\tau))_{\tau=t'}^{t}$. To denote the extension of a transcript by a single round  or the concatenation of two transcripts, we use $\oplus$: for example, we may write
\begin{align*}
    \pi_{1:t} &= \pi_{1:t-1} \oplus \pi_t\\
    \pi_{1:T} &= \pi_{1:t} \oplus \pi_{t+1:T}.   
\end{align*}
We write $\Pi^*=(\cX \times [0,1] \times [0,1])^*$ as the domain of all transcripts. Formally, the adversary is modelled as a probabilistic mapping $\mathrm{Adv}:\Pi^*\rightarrow \Delta(\cX \times [0,1])$ from transcripts to distributions over data points and conformal scores. The learner is modeled as a mapping $\mathrm{Learn}:\Pi^*\rightarrow (\cX \rightarrow \Delta [0,1])$ from transcripts to a probabilistic mapping from feature vectors $x$ to distributions over $[0,1]$ (distributions over conformity thresholds). 
Fixing both a learner and an adversary induces a probability distribution over transcripts. Our goal is to derive algorithms that have probabilistic guarantees over the randomness of the transcript distribution, in the worst case over all possible adversaries.

Given a transcript $\pi_{1:T}$, a group $G \in \cG$ and a set of rounds $S \subseteq [T]$, we write 
\[
    G_S = \{ t \in S: x_t \in G\}.
\]
In words, this is the set of rounds in $S$ in which the realized feature vectors in the transcript belonged to $G$.  When it is clear from context, we sometimes overload the notation, and for a group $G \in \cG$, and a period $t \leq T$, write $G_t$ to denote the set of data points (indexed by their rounds) in a transcript $\pi_{1:t}$ that are members of the group $G$:
\[
    G_t = \{\tau \in [t]: x_\tau \in G\}.
\]

Given some threshold $q$, we say the conformity threshold covers the conformal score $s$ if 
\begin{align*}
    \cover(q, s)\equiv \ind[s \le q]=1.
\end{align*}

For any $S \subseteq [T]$ we define
\[  
    \bH(S) = \frac{1}{|S|} \sum_{t \in S} \cover\left(q_t, s_t \right)
\]
to denote the empirical coverage rate over the set of rounds $S$.

To define threshold calibration, we bucket our thresholds using a discretization parameter $m$. For any $m$ and bucket index $i \in [m-1]$, we write $\Bm(i) = \left[\frac{i-1}{m}, \frac{i}{m}\right)$ and $\Bm(m) = \left[\frac{n-1}{m}, 1 \right]$ so that these buckets evenly partition the unit interval $[0, 1]$\footnote{We can handle non-uniform discretization of the unit interval as well without any additional complication}. Conversely, given a threshold $q \in [0, 1]$, we write $\Bminv(q) \in [m]$ for the index of the bucket that $w$ belongs to: $\Bminv(q) = i$, for $i$ such that $q \in \Bm(i)$.  When clear from the context, we elide the subscript $n$ and write $B(i)$ and $\Binv(q)$.

For any $S \subseteq [T]$ and $i \in [n]$, we define $S(i)$ to be the subset of rounds in $S$ in which the learner's threshold falls in bucket $i$. Formally,
\[
    S(i) = \left\{t \in S: q_t \in \Bm(i) \right\}.
\]

We can now define threshold calibrated multivalid coverage.
\begin{definition}[Threshold Calibrated Multivalid Coverage]
\label{def:multivalid}
Fix a coverage target $1-\delta$ and a collection of groups $\cG \subset 2^\cX$. Given a transcript $\pi_{1:T}$, a sequence of conformity thresholds $(q_t)_{t=1}^T$ is said to be \emph{$(\alpha, m)$-multivalid with respect to $\delta$ and $\cG$} for some function $\alpha:\bN \to \R$, if for every $i \in [m]$ and $G \in \cG$, the following holds true:
\[
     \left|\bH(G_T(i)) - (1-\delta) \right| \le \alpha(|G_t(i)|) 
\]
\end{definition}
Note that multivalid coverage is defined by a \emph{function} $\alpha$ of the length of the sequence on which we are computing empirical coverage. This allows us to give fine-grained bounds that scale with the length of this sequence. Throughout this paper we use following family of functions, parameterized by a constant $\epsilon > 0$:
\begin{align*}
    \alpha(n) = \frac{f(n)}{n} \quad\text{and}\quad f(n) = \sqrt{(n+1) \log^{1+\epsilon}(n+2)}
\end{align*}
A useful fact is that the series $\frac{1}{f(n)^2}$ is convergent:
    $\sum_{n=0}^\infty \frac{1}{f(n)^2} = K_\epsilon$
where $K_\epsilon$ is a constant depending only on our choice of $\epsilon$ that will appear in our bounds.
\begin{remark}
We note that obtaining coverage at the rate of $\alpha(n) = \tilde O\left(\frac{1}{\sqrt{n}} \right)$ on subsequences of length $n$ is the best rate possible for threshold calibrated coverage: even if the labels were drawn from a \emph{known} distribution rather than being selected by an adversary, and even if we produced prediction sets with exactly the correct  coverage rate over the distribution, we would expect that our empirical coverage on a sequence of length $n$ would differ from our expected coverage at this rate.
\end{remark}

\section{Our Algorithm and Analysis}
\label{sec:anal}
Before we provide the algorithm and its guarantees, we first discuss a  needed assumption. Observe that even in the easier distributional setting where the conformal score $s$ is drawn from a fixed, known distribution: $s \sim \cD$ --- there may not be any threshold $q \in [0, 1]$ that satisfies the desired target coverage value, i.e. that guarantees that $\left\vert \E_{s \sim \cD}[\cover(q, s) - (1-\delta)] \right\vert$ is small. Consider for example a distribution that places all its mass on a single value $s$. Then any threshold $q$ covers the $s$ with probability $1$ or probability $0$, which for $\delta \not\in \{0,1\}$ is bounded away from our target coverage probability. One could randomize the threshold to get the target marginal coverage rate, but this corresponds to the ``cheating'' strategy we outline in footnote 1, and in particular would not satisfy our notion of \emph{threshold calibrated} coverage. Of course, if achieving the target threshold-calibrated coverage is impossible in the easier distributional setting, then it is also impossible in the more challenging online adversarial setting. 

With this in mind, just as with many other approaches to conformal prediction that aim to converge to the correct coverage rate (rather than conservatively over-cover), we will need to assume that our target distributions are not too concentrated on any single point. Following \cite{multivalid}, we define a class of smooth distributions for which achieving (approximately) the target coverage is always possible for some threshold $q$  defined over an appropriately finely discretized range. Our smoothness condition makes sense even for discrete distribution, so we do not need to assume continuity. To denote the uniform grid on $[0,1]$, we write
\[
    \cP^{rn} = \left\{ 0, \frac{1}{rm}, \frac{2}{rm}, \ldots, 1\right\}. 
\]
We show that we can achieve (approximately) our target coverage goals in the online adversarial setting when the adversary is constrained to playing smooth distributions, which are distributions that do not put too much probability mass on any sufficiently small sub-interval. 

\begin{definition}
\label{def:smoothness}
A distribution $Q \in \Delta([0, 1])$ is $(\rho,rm)$-\emph{smooth} if for any $0 \leq a \leq b \leq 1$ such that $|a-b| \leq \frac{1}{rm}$,
\[
\Pr_{s \sim Q}[s \in [a,b]] \leq \rho.
\]
We say that a joint distribution $\cD \in \Delta (\cX \times [0, 1])$ is $(\rho, rm)$-\emph{smooth} if for every $x \in \cX$, the marginal conformal score conditional on $x$,  $\cD\vert_{x}$, is $(\rho, rm)$-smooth. We say an adversary is $(\rho,rm)$-smooth if the joint distribution over $(x_t,s_t)$ is $(\rho, rm)$-smooth at every round $t\in[T]$.
\end{definition}

\begin{obs}
\label{obs:smooth}
For any $\delta \in [0,1]$, fixed ($\rho,rm$)-smooth score distribution $Q \in \Delta [0,1]$, there always exists some threshold $q \in \cP^{rm}$ such that
\[
\left\vert \Pr_{s \sim Q}[\cover(q, s)] - (1-\delta) \right\vert \le \rho.
\]
\end{obs}
\begin{remark}
For any $\rho$, the assumption of $(\rho,rm)$-smoothness becomes more mild as $r \rightarrow \infty$.  For us, $r$ will be a nuisance parameter that we can choose to be as large as we want --- we will not have to pay for it either in our running time or our coverage bounds. We can also \emph{algorithmically enforce} smoothness by perturbing the conformal scores with small amounts of noise from any continuous distribution, and so we should think of smoothness as a mild assumption. Our experiments bear this out. 
\end{remark}

We now present the algorithm (MVP --- MultiValid Predictor)  along with its guarantees and provide the analysis in Section \ref{subsec:alg-analysis}. It resembles the algorithm for online mean multicalibration given in \cite{multivalid}, which in turn is a multi-group generalization of the ``almost deterministic'' calibration algorithm of \cite{foster2021forecast}.

\begin{algorithm}[H]
\SetAlgoLined
\begin{algorithmic}
\FOR{$t=1, \dots, T$}
    \STATE Take as input an arbitrary conformal score $s_t:\cX \times \cY\rightarrow [0,1]$
	\STATE Observe $x_t$ and for each $i \in [m]$ and $G \in \cG(x_t)$, compute 
	\begin{align*}
	    n^{G,i}_{t-1} &= \left(\sum_{\tau=1}^{t-1} \ind[q_\tau \in B(i),~ x_\tau \in G]\right) & \textrm{Definition \ref{def:group-bucket-size}} \\
	    V^{G,i}_{t-1} &= \sum_{\tau=1}^{t-1} \ind[x_\tau \in G, q_\tau \in \Bm(i)] \cdot 
	    \left(\cover(q_\tau, s_\tau) - (1-\delta)\right) & \textrm{Definition \ref{def:coverageerror}} \\
	    C^i_{t-1}(x_{t}) &= \sum_{G \in \cG(x_t)} \frac{\exp\left(\eta \frac{V_t^{G,i}}{f(n_t^{G,i})}\right) - \exp\left(-\eta \frac{V_t^{G,i}}{f(n_t^{G,i})}\right)}{f(n_t^{G,i})}. & \textrm{From Lemma \ref{lem:surrogate-loss-increase}}
	\end{align*}
    \IF {$C^i_{t-1}(x_{t}) > 0$ for all $i \in [m]$}
        \STATE Output $q_t= 0$.
    \ELSIF {$C^i_{t-1}(x_{t}) < 0$ for all $i \in [m]$}
        \STATE Output $q_t = 1$.  
    \ELSE 
        \STATE Find $i^* \in [m-1]$ such that $C^{i^*}_{t-1}(x_{t}) \cdot C^{i^*+1}_{t-1}(x_{t}) \leq 0$
        \STATE Define $0 \leq p_t \leq 1$ as follows (using the convention that 0/0 = 1):
        \[
        p_t = \left|C^{i^*+1}_{t-1}(x_{t}) \right| / \left(|C^{i^*+1}_{t-1}(x_{t})| + |C^{i^*}_{t-1}(x_{t})|\right).
        \]
        \STATE Choose threshold $q_t =   \frac{i^*}{m}- \frac{1}{rm}$ with probability $p_t$ and $q_t = \frac{i^*}{m}$ with probability $1-p_t$.
    \ENDIF
    \STATE Output prediction set $\cT_t(x_t) = \{y \in \cY : s_t(x_t,y) \leq q_t\}$
\ENDFOR
\end{algorithmic}
\caption{MVP($\eta,n,r$)}
\label{alg:conformity-score-multivalidator}
\end{algorithm}

\begin{restatable}{theorem}{thmalgmutivalidguarantee}
\label{thm:alg-mutivalid-guarantee}
Set $\eta =\sqrt{\frac{\ln(|\cG|m)}{2K_\epsilon |\cG|m}}$. Against any $(\rho,rm)$-smooth adversary and for any adaptively chosen sequence of conformal scores $s_t$, MVP (Algorithm~\ref{alg:conformity-score-multivalidator}) produces a sequence $(q_t)_{t=1}^T$ that is $(c_{\text{exp}} \alpha (\cdot),m)$-multivalid in expectation over the randomness of $\pi_{1:T}$ with respect to $\delta$ and $\cG$ where 
    \[
        c_{\text{exp}} \le \sqrt{4K_\epsilon |\cG| m \ln(|\cG| m)} + \rho T.
    \]
\end{restatable}

\begin{remark}
Since we can take $r$ to be arbitrarily large, for any continuous distribution we can drive the $\rho T$ term to zero. Thus this bound establishes nearly statistically optimal convergence rates for constant $|\cG|$ and $m$. Using a simpler analysis analogous to that of \cite{multivalid} for mean multicalibration, and an ``un-normalized score function''  $C^i_{t-1}(x_{t}) = \sum_{G \in \cG(x_t)} \exp\left(\eta V_t^{G,i}\right) - \exp\left(-\eta V_t^{G,i}\right)$, it is also possible to establish $(\alpha, m)$-multivalidity with $\alpha(n) = O(\sqrt{T\log(|\cG| m)}/n + \rho)$, which has an optimal dependence on $|\cG|$ and $m$, but has a bad dependence on $T$. We believe that our sub-optimal dependence on $|\cG|$ and $m$ is an artifact of our analysis, and not a property of our algorithm. We implement both the normalized and un-normalized version of the algorithm in our codebase, and find that both perform comparably.
\end{remark}

\subsection{Analysis}
\label{subsec:alg-analysis}
In this section we outline the analysis of our algorithm. Full proofs are in Appendix \ref{app:anal}. 

For each group $G \in \cG$, bucket $i \in [m]$  time $t \in [T]$, we'd like to bound the coverage error on the subsequence of rounds $\tau$ in which $x_\tau \in G$ and $q_\tau \in B_m(i)$ in terms of the length of that sequence. We give the following notation for these sequence lengths:
\begin{definition}[Group-bucket size]
\label{def:group-bucket-size}
Given a transcript $\pi_{1:t} = ((x_{\tau}, s_{\tau}, q_{\tau}))_{\tau = 1}^t$, we define the size for a group $G \in \mathcal{G}$ and a bucket $i \in [m]$ at time $t$ to be:
\[
    n_t^{G,i}(\pi_{1:t}) = \left(\sum_{\tau=1}^t \ind[q_\tau \in B(i),~ x_\tau \in G]\right).
\]
When the transcript is clear from context, we will sometimes write $n^{G,i}_t$.
\end{definition}

Similarly, for each $G \in \cG$, $i \in [m]$ and time $t \in [T]$, we can define the (un-normalized) coverage error on the sequence corresponding rounds $\tau \leq t$ such that $x_\tau \in G$ and $q_\tau \in [m]$:

\begin{definition}
\label{def:coverageerror}
Given a transcript $\pi_{1:t} = ((x_\tau, s_\tau, q_\tau))_{\tau=1}^t$,  we define the coverage  error for a group $G \in \cG$ and bucket $i \in [ m]$ at time $t$ to be:
\begin{align*}
&V_t^{G,i} = \sum_{\tau=1}^t \ind[x_\tau \in G, q_\tau \in \Bm(i)] \cdot  v_\delta(q_\tau, s_\tau)
\end{align*}
where $v_\delta(q, s) = \cover(q, s) - (1-\delta)$.
\end{definition}
Note that $V_t^{G,i}$ just records the deviation of the empirical coverage from its target $(1-\delta)$ on the subsequence of rounds $\tau$ in which $x_\tau \in G$ and $q_t \in B_m(i)$: it takes a positive value if we have \emph{over-covered} on this subsequence and a negative value if we have under-covered.

\begin{obs} 
\label{obs:multicoverage}
Fix a transcript $\pi_{1:T}$. If for all $G \in \cG$ and $i \in [m]$, we have that
\[
\frac{\left|V_T^{G,i}\right|}{f(n^{G,i}_T)} \leq c
\]
for some constant $c$, then the corresponding sequence of thresholds widths $(q_t)_{t=1}^T$ is $(c \alpha(\cdot), m)$-multivalid with respect to $\delta$ and $\cG$. 
\end{obs}

Thus the end-goal of our analysis is to uniformly upper bound $\frac{\left|V_T^{G,i}\right|}{f(n^{G,i}_T)}$ over all $G \in \cG$ and $i \in [m]$. The analysis in this section can be seen as an extension of the surrogate loss argument developed in \cite{multivalid} for the problem of mean multicalibration. There are two main novel insights that lead to our algorithm and analysis for multivalid coverage. \cite{multivalid} were unable to extend their simple multicalibration algorithm to prediction interval multivalidity (and instead analyzed an impractical Ellipsoid-based algorithm). Informally this is because they parameterized prediction intervals with two parameters (the lower and upper endpoint), which eliminated the simple one-dimensional structure they were able to exploit for multicalibration. In contrast, our prediction intervals are parameterized by a single parameter $q$, which allows us to exploit the one-dimensional structure that allows us to derive a simple, combinatorial algorithm that is similar to the multicalibration algorithm from \cite{multivalid}. Second, the bounds in \cite{multivalid} uniformly bound the coverage error for each group $G \in \cG$ and bucket $i \in [m]$ by $\tilde O(\sqrt{T})$, which is optimal only for subsequences that have $n_T^{G,i} = \Omega(T)$. In contrast, we obtain non-uniform bounds that are independent of $T$ and depend only on $n_T^{G,i}$, and (at least for constant $m$ and $|\cG|)$ have the optimal $\sqrt{n_T^{G,i}}$ dependence. We do this by analyzing a modified surrogate loss function. This leads to a significant amount of added complexity which accounts for the bulk of our argument.

To bound the maximum absolute value of our coverage errors divided by $f(n^{G,i}_T)$ across all groups and buckets  (i.e. $\max_{G \in \cG, i \in [m]} \frac{V^{G,i}_T}{f(n^{G,i}_T)}$), we use the following surrogate loss functions: 
\begin{definition}[Surrogate loss]
Fix a transcript $\pi_{1:t} \in \Pi^*$ and a parameter $\eta \in (0,1/2)$. Define a surrogate coverage loss function at day $t$ for bucket $i \in [n]$ and group $G \in \cG$ as
\[
L_t^{G,i}(\pi_{1:t}) = \left(\exp\left(\eta \frac{V_t^{G,i}}{f(n_s^{G,i})}\right) + \exp\left(-\eta \frac{V_t^{G,i}}{f(n_s^{G,i})}\right) \right)
\]
where $V_t^{G,i}$ are implicitly functions of $\pi_{1:t}$. Similarly, we denote the overall surrogate coverage loss function as
\[
L_t(\pi_{1:t}) = \sum_{\substack{G \in \cG,\\i \in [m]}} L^{G,i}_t(\pi_{1:t}).
\]
 When the transcript is clear from context we will sometimes simply write $L^{G,i}_t$ and $L_t$. 
\end{definition}

Our strategy will be to prove that our algorithm guarantees that the surrogate loss is small, which will then allow us to conclude that our coverage error is small for every group $G \in \cG$ and bucket $i \in [m]$. We first show that  the increase in the surrogate loss can be bounded in the following way:
\begin{restatable}{lemma}{lemsurrogatelossincrease}
\label{lem:surrogate-loss-increase}
Fix $\eta \in (0,\frac{1}{2})$ and a transcript $\pi_{1:t-1}$ for some round $t-1$. Then, for any $\pi_t = (q_t, x_t, s_t)$, we have
\[
    L_t(\pi_{1:t-1} \oplus \pi_t) - L_{t-1}(\pi_{1:t-1}) \le \sum_{(G,i) \in A_t(\pi_t)} \eta v_{\delta}(q_{t}, s_{t}) C_{t-1}^{G,i} + \frac{2\eta^2}{f(n_t^{G,i})^2} L_{t-1}^{G, i}(\pi_{t-1}) 
\]
where for any round $t \in [T]$,
\[
    A_t(\pi_t) = \{(G,i): G \in \cG(x_t), i = \Binv(q_t)\}
\]
is the set of $(G,i)$ pairs that are ``active'' in round $t \in [T]$
and
\begin{equation*}
    C_t^{G, i} = \frac{\exp\left(\eta \frac{V_t^{G,i}}{f(n_t^{G,i})}\right) - \exp\left(-\eta \frac{V_t^{G,i}}{f(n_t^{G,i})}\right)}{f(n_t^{G,i})}.
\end{equation*}
\end{restatable}

Next, we show that Algorithm~\ref{alg:conformity-score-multivalidator} guarantees that the first term in the surrogate loss increase  $\sum_{(G,i) \in A_t(\pi_t)} v_{\delta}(q_{t}, s_{t}) C_{t-1}^{G,i}$ is small in  expectation over the randomness of the algorithm, whenever the score distribution is smooth.
\begin{restatable}{lemma}{lemalgboundedincreasesurrogateloss}
\label{lem:alg-bounded-increase-surrogate-loss}
Fix any $t \in [T]$, $\eta \in (0,\frac{1}{2})$, transcript $\pi_{1:t-1}$ recording a realization for the first $t-1$ rounds and $x_t$. At round $t$,   Algorithm~\ref{alg:conformity-score-multivalidator} chooses a distribution over $q_t$ such that against any $(\rho,rm)$-smooth distribution over conformal scores $s_t$, we have:
\begin{align*}
\E_{(s_t, q_t)}\left[\sum_{(G,i) \in A_t(\pi_t)} v_{\delta}(q_{t}, s_{t}) C_{t-1}^{G,i} \middle| \pi_{1:t-1} \right] \le \rho L_{t-1}
\end{align*}
\end{restatable}

Carefully telescoping the bounded increase in surrogate loss over each round via Lemma~\ref{lem:surrogate-loss-increase} and \ref{lem:alg-bounded-increase-surrogate-loss} and using the fact that $\{1/f(n)^2\}_{n=1}^\infty$ is a convergent series yields Theorem~\ref{thm:alg-mutivalid-guarantee}. We carry out the argument in detail in the appendix.

\section{Experiments}
\label{app:experiments}
In this section, we evaluate MVP and compare it to more traditional methods of conformal prediction on a variety of tasks. In each comparison, we use the same model and conformal score for MVP and for the methods we compare against --- the only difference is the type of the conformal prediction wrapper. Our code is available at \href{https://github.com/ProgBelarus/MultiValidPrediction }{https://github.com/ProgBelarus/MultiValidPrediction}.

First in Section \ref{sec:exp-exchange} we study a synthetic regression problem in a simple exchangeable (i.i.d.)  setting, and compare to split conformal prediction \citep{lei2018distribution}. We show that even when we measure only marginal empirical coverage, MVP improves over split conformal prediction when the regression function must be learned. This is because to maintain the exchangeability of conformal scores, split conformal prediction must split the data into two sets --- one for training the regression function and one for calibrating the prediction sets.\footnote{This is not only a theoretical requirement --- split conformal prediction fails badly otherwise.} In contrast, our method does not require exchangeability, so we can both train the regression model and calibrate our prediction sets on the entire dataset. Then, we modify our regression problem so that there are 20 overlapping sub-populations, and one of the sub-populations (consisting of half of the data points) has higher label noise. We measure group-wise coverage for MVP, for naive split conformal prediction that has no knowledge of the groups to be covered, and the method of \cite{barber2019limits} which guarantees (conservative) group-wise coverage for intersecting groups. We find that MVP significantly improves on both methods. Finally we run all three of these methods on real data drawn i.i.d. from a U.S. Census dataset provided by the Folktables package \citep{ding2021retiring}, where we ask for group-wise coverage on groups defined by race and sex designations. Again, we find that MVP consistently obtains the closest to its target group-wise coverage while providing narrower prediction intervals. 

Next, in Section \ref{sec:exp-shift} we study a regression problem in the presence of covariate shift. First we replicate an experiment of \cite{tibshirani2019conformal}, in which a synthetic covariate shift (with known propensity scores and known changepoint) is simulated on a UCI dataset. The method of \cite{tibshirani2019conformal} reweights the calibration set using the propensity scores. MVP can also take advantage of propensity scores when they are known: we  give MVP a ``warm start'' from the same portion of the dataset that split conformal prediction uses for calibration, sampled with replacement after being re-weighted by the propensity scores. Both algorithms are then evaluated on the shifted distribution. We find both algorithms perform comparably. We then experiment with unknown and unanticipated covariate shift simulated on datasets derived by U.S. Census data provided from the Folktables package \citep{ding2021retiring}: We compare to split conformal prediction calibrated on the California data (this time without re-weighting) and evaluated on the Pennsylvania data. Similarly, we again give MVP a warm start on the California data (again without reweighting), and then measure its performance on 2018 Pennsylvania Census data.  We find that MVP  obtains the correct coverage rate and smaller interval widths compared to the split conformal method despite having no knowledge of the distribution shift.

In Section \ref{sec:exp-time} we evaluate MVP on time series data --- 20 years of stock returns, in a volatility prediction task. We compare MVP to the Adaptive Conformal Inference (ACI) method of \cite{gibbs2021adaptive}, which guarantees marginal (but not threshold calibrated) coverage for adaptively chosen data. When evaluated in terms of marginal coverage, we find that MVP and ACI perform comparably: ACI obtains average coverage slightly closer to the target, whereas MVP predicts a more stable sequence of thresholds. We then complicate the experiment to exhibit the two advantages of MVP (groupwise coverage and threshold calibrated coverage). First we define 20 intersecting groups defined as the trading days that are multiples of $1,2,\ldots,20$. We add perturbations to the stock returns that differ across these groups, and find that MVP continues to produce the correct group-wise coverage, whereas ACI fails to. Next, we produce a fully adversarial sequence by presenting examples to the algorithms not in time order but in \emph{sorted order by their conformal scores}. By construction, this sequence would cause split conformal prediction methods to have 0 coverage, but both ACI and MVP are required to obtain the correct marginal coverage on this sequence. However, we find that given this sequence, ACI reduces to a strategy that, similar to the uninformative ``cheating'' strategy mentioned in Footnote 1, predicts the trivial coverage interval (all of $[0, 1]$) on most days --- which guarantees marginal but not threshold calibrated coverage, and does not produce non-trivial average interval widths. In contrast, MVP, by virtue of its threshold calibration condition, produces a sequence of coverage thresholds that correctly track the sequence of conformal scores of the true labels in the data, and hence produces prediction intervals with the correct widths.

Finally in Section \ref{sec:imagenet} we compare MVP to the work of \cite{angelopoulos2020uncertainty} on a large-scale ImageNet classification task. We find that MVP obtains comparable coverage rates and prediction set sizes, despite the fact that the setting is favorable to \cite{angelopoulos2020uncertainty} --- i.e. the data is i.i.d. and we measure only marginal coverage.

\subsection{Exchangeable Data}
\label{sec:exp-exchange}

\subsubsection{Basic Experimental Setup and Marginal Coverage}
\label{sec:sub-marginal}
We simulate a synthetic linear regression problem in which the regression model must be trained in tandem with the conformal predictor. The feature domain consists of 10 binary features and 290 continuous features. For any input $x$, the binary features are drawn from a uniform distribution and each continuous feature is drawn from a normal distribution $\mathcal{N}(0, \sigma_x^2)$. Each example's label is governed by an ordinary least squares model:
\begin{equation*}
    y = \langle \theta, x \rangle + \mathcal{N}(0, \sigma_y^2)
\end{equation*}
for some fixed vector $\theta \in \R^{300}$ unknown to the learner.

 We run both MVP and split conformal prediction \citep{lei2018distribution} using the conformal score $s_t(x, y) = |f_t(x) - y|$. When running MVP, we train $f_t$ using least squares regression on all points $(x_{t'},y_{t'})$ for $t' < t$. For split conformal prediction, we divide points evenly between a calibration set and a training set (points from odd time steps go into the calibration set, points from even time steps go into the training set), and $f_t$ is trained using least squares regression on all points in the training set at time $t-1$. (We also tried training $f_t$ on all points, but this causes split conformal prediction to fail catastrophically). 

\begin{figure}
\includegraphics[width=0.5\linewidth]{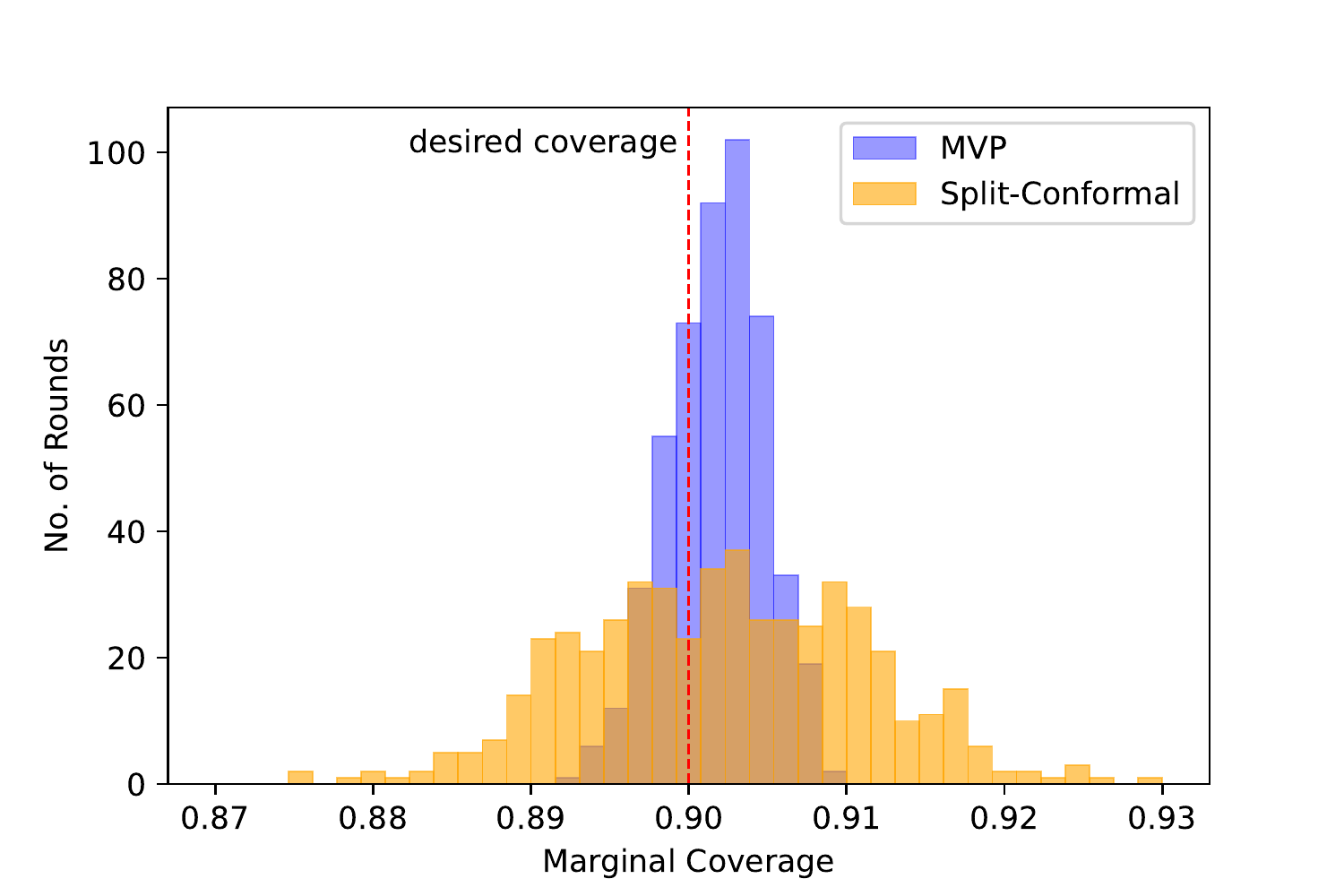}
\includegraphics[width=0.5\linewidth]{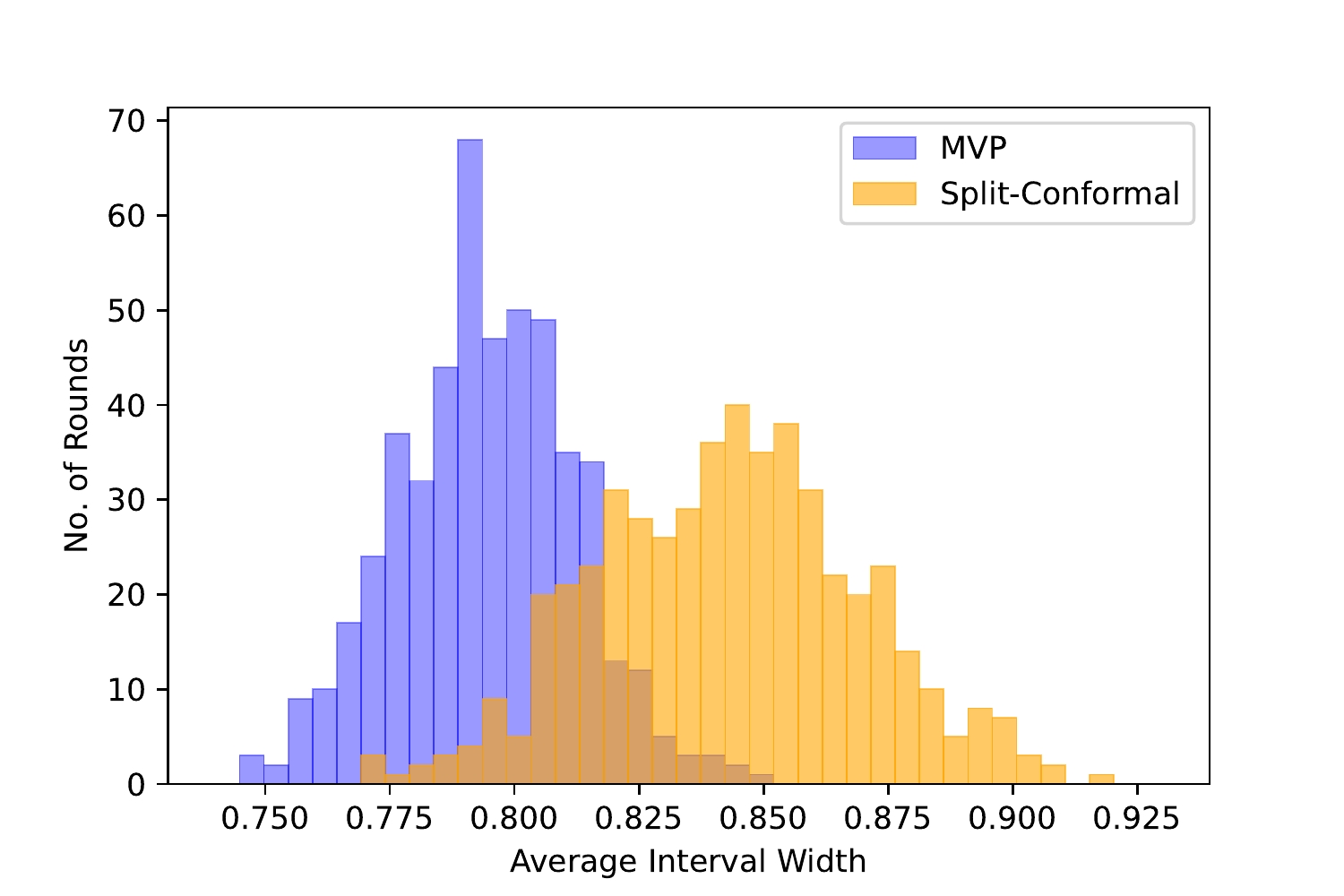}
\caption{The plot on the left is a histogram of the empirical marginal coverage of MVP and split conformal prediction over 500 repeated trials; the right hand plot is similarly a histogram of the average interval width for both methods. We see that MVP gets both empirical coverage that is more tightly concentrated around its target (0.9) and narrower coverage interval width.}
\label{fig:marginal}
\end{figure}

\paragraph{Results} We set $\sigma_x^2 = 0.1$, $\sigma_y^2 = 0.2$ and run 500 independent trials of our experiment, each for $T = 2000$ steps. $\theta$ is independently selected for each trial. The results are shown in Figure \ref{fig:marginal}. MVP simultaneously obtains empirical coverage that is more tightly concentrated around its target and obtains narrower coverage intervals compared to split conformal prediction. Despite the fact that we are in a setting that is extremely favorable to split conformal prediction (i.i.d. data and marginal coverage evaluation), MVP has the advantage that it can use a regression function $f_t$ trained on \emph{all} past data, without the need to set aside a calibration set. This is needed for split conformal prediction to maintain the exchangeability of the conformal scores.

\subsubsection{Multi-Group Coverage}
\label{sec:sub-marginal-2}
We now compare the coverage of MVP to that of split conformal prediction not just marginally, but group-wise. We use the same feature generation process and conformal score as for our marginal coverage experiment described in Section \ref{sec:sub-marginal}. Recall that the first 10 of the 300 features in our data domain are binary, which we now use to define 20 (intersecting) groups defined by the value of each of the 10 binary features. Labels are still generated according to an ordinary least squares model, but now the noise rate depends on the groups that each datapoint is a member of. Specifically:
\begin{equation*}
    y = \langle \theta, x \rangle + \mathcal{N}\left(0, \sigma^2 + \sum_{i=1}^{10} \sigma_i^2 x_i\right)
\end{equation*}
for some fixed vector $\theta \in \mathbb{R}^{300}$, and for fixed values of $\sigma_i$, each associated with one of the binary features indicating groups. 

We run MVP parameterized to promise multi-valid coverage for the set of 20 intersecting groups defined by the first 10 binary features of the input: For each $i \in \{0, 1, \cdots, 19\},$ we define $G_i = \{x \in \mathcal{X} \mid x_{\lceil (i + 1)/2\rceil} \equiv_2 i \}$
and let $\mathcal{G} = \{G_i \mid 0 \leq i \leq 19\}$. At each time-step $t$, we train a regression model $f_t$ on all past data.

We compare to two benchmark conformal prediction methods. First, we compare to naive split conformal prediction (which ignores the group structure), just as in Section \ref{sec:sub-marginal}. This method offers no guarantees about group-wise coverage. Second, we compare to the method of \cite{barber2019limits} which separately computes a calibration threshold for each of the 20 groups marginally, and on each example $x_t$, uses the most conservative (i.e. largest) threshold associated with any of the groups for which $x_t$ is a member. This method guarantees coverage \emph{at least} the target coverage level, but does not guarantee coverage approaching the target.  Note there are $2^{10}$ different subsets of groups that each example might be a member of, and so the method of \cite{romano2020malice} which separately calibrates on \emph{disjoint} groupings of the data cannot be run without having roughly 1000-fold more data. For both conformal prediction methods we equally split the data between a training set used for training the regression model $f_t$ and a calibration set used for calibrating the prediction intervals. We run MVP with $m=40$ calibration buckets.

\begin{figure}
\includegraphics[width=0.5\linewidth]{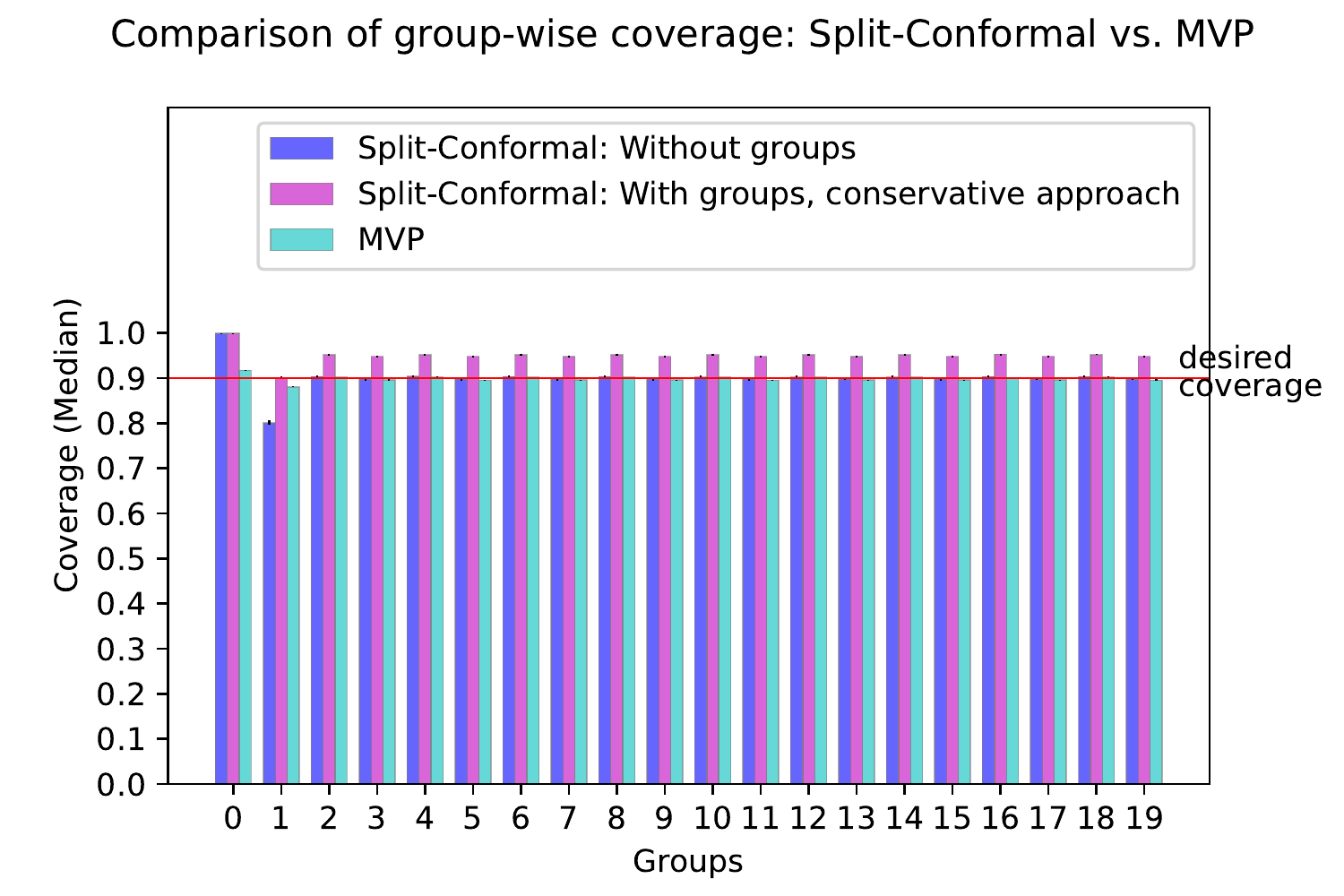} 
\includegraphics[width=0.5\linewidth]{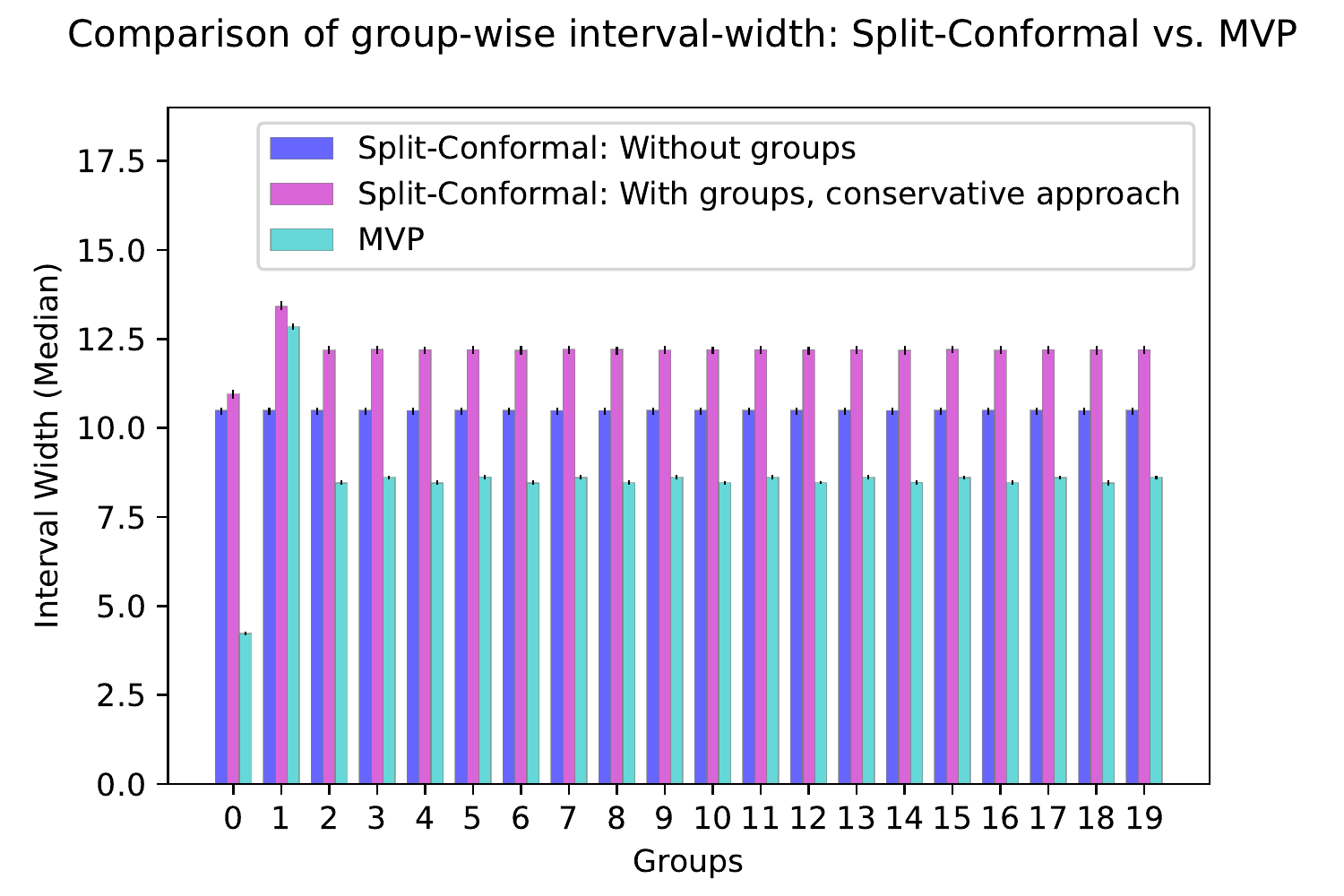}
\caption{On the left we plot the median over 100 independent trials of the coverage conditional on membership in each of our 20 groups. On the right we plot the median of the average interval width conditional on membership in each of the 20 groups. Compared to the split conformal prediction methods, we see that MVP obtains the target coverage level on each group (neither under nor over covering), and obtains narrower interval widths. 
The error bars represent 25th and 75th quantiles, and they are not easily visible in this figure as they are quite narrow: for conformal with groups, both bar endpoints are within $\pm 0.0039$ from the median,
for conformal without groups, within $\pm 0.0054$ from the median, and for MVP, within $\pm 0.0021$ from the median.}
\label{fig:groupwise}
\end{figure}

\paragraph{Results} We run 100 independent trials of our experiment, each for $T = 20,000$ data points. Our results are plotted in Figure \ref{fig:groupwise}. We set $\sigma_1^2 = 3.0$  and $\sigma_2^2 = \ldots = \sigma_{10}^2 = 0.1$ so that $G_0$ is a ``low noise'' group and $G_1$ is a ``high noise'' group. We keep the values of the $\sigma_i$ fixed across all trials, but each is run with an independently drawn $\theta$. As expected, we find that naive split conformal prediction fails to meet its coverage target, over-covering on the low noise group and under-covering on the high noise group, and uses a uniform interval width. In contrast, both MVP and the conservative method of \cite{barber2019limits} use different average interval widths for different groups. The conservative method of \cite{barber2019limits} always gets at least the target coverage, but significantly over-covers on every group except for the high noise group. In contrast MVP obtains the target coverage on every group. MVP also has lower average interval width on every group compared to \cite{barber2019limits}, and (correctly) produces  significantly narrower intervals on the low noise group. 

\subsubsection{Multi-Group Coverage with Folktables Data}
We now evaluate the group-wise performance of MVP against the same two split conformal prediction methods on a real dataset derived from the 2018 Census American Community Survey Public Use Microdata provided by the Folktables package  \citep{ding2021retiring}. The dataset includes instances of people from all the states in the USA; for this experiment, we consider only those instances from the state of California. There are 195665 instances of this kind, and we subsample this data (0.1 for training, 0.1 for testing). 

Our goal in this experiment is to generate prediction sets for a person's income. The Folktables dataset has nine different codes for race\footnote{1. White alone, 2. Black or African American alone, 3. American Indian alone, 4. Alaska Native alone, 5. American Indian and Alaska Native tribes specified; or American Indian or Alaska Native, not specified and no other races, 6. Asian alone, 7. Native Hawaiian and other Pacific Islander alone, 8. Some Other Race alone, 9. Two or More Races.}  and two codes for sex\footnote{1. Male, 2. Female.}. Note that the race and sex groups intersect.  We define groups for five out of nine of the race  groups (the remaining  four have very little data) and groups for both sexes, for a total of seven groups. We run MVP with $m = 40$ buckets, and parametrized to promise multi-valid coverage for each of these seven groups, and compare against both conformal prediction methods introduced in Section \ref{sec:sub-marginal-2}. 

Using the training data, we train a linear regression model $f$ to predict income and use it to define the conformal score $s(x, y) = |f(x) - y|$ for all three methods. An initial calibration set of size 1000 (taken from the test data) is used for both split conformal methods, and is used as a "warm start" for MVP (i.e. this data is used to update variables used in the algorithm, but we do not record performance over these instances). The remaining test data is used to compare performance between methods. For the split conformal methods, the calibration set grows to include the previously observed examples from the test set as time goes on.

\begin{figure}
\includegraphics[width=0.93\linewidth]{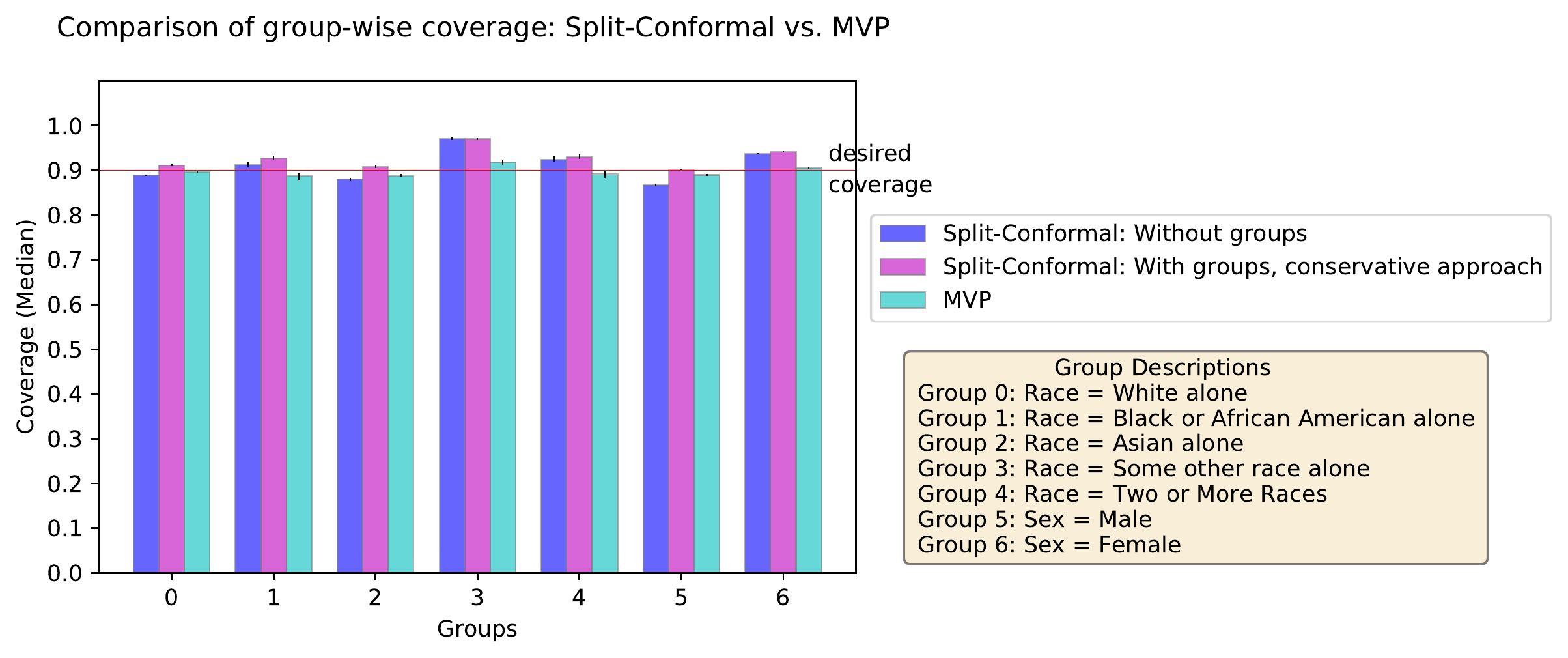} \\ 
\includegraphics[width=0.93\linewidth]{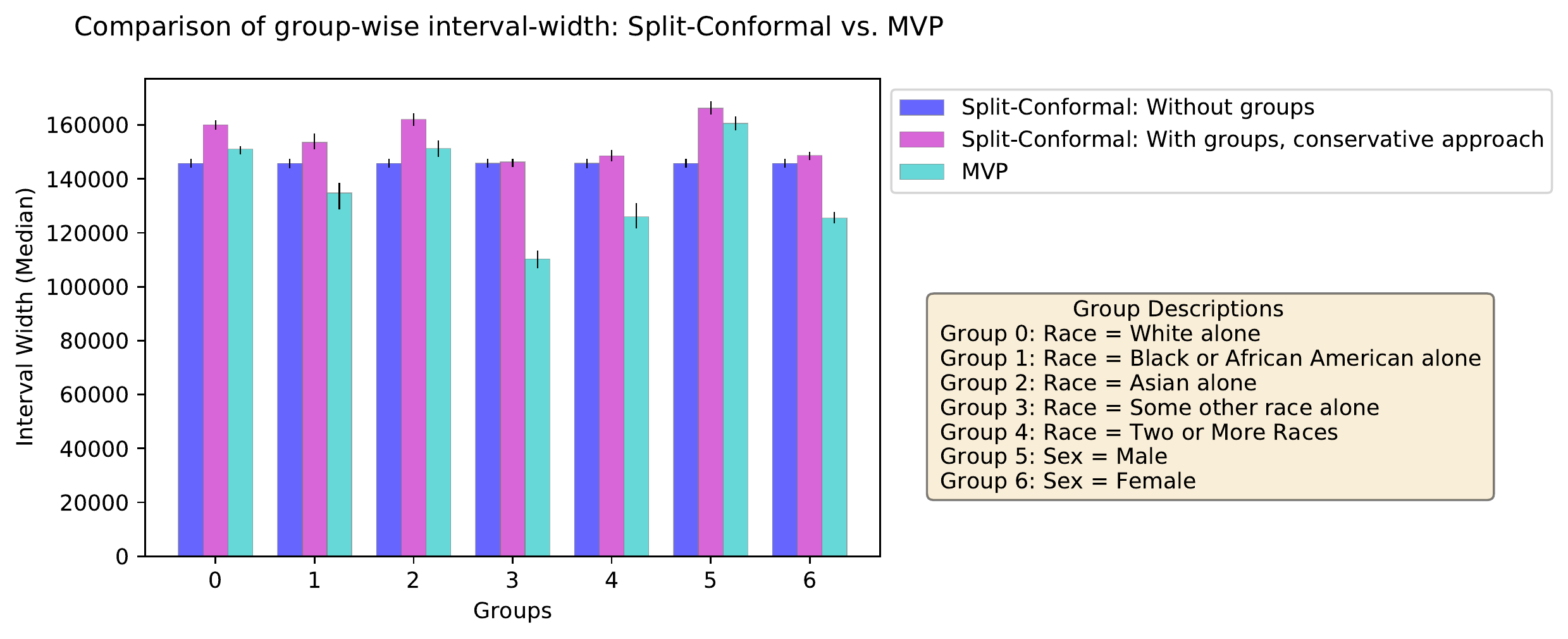} \\ 
\includegraphics[width=0.93\linewidth]{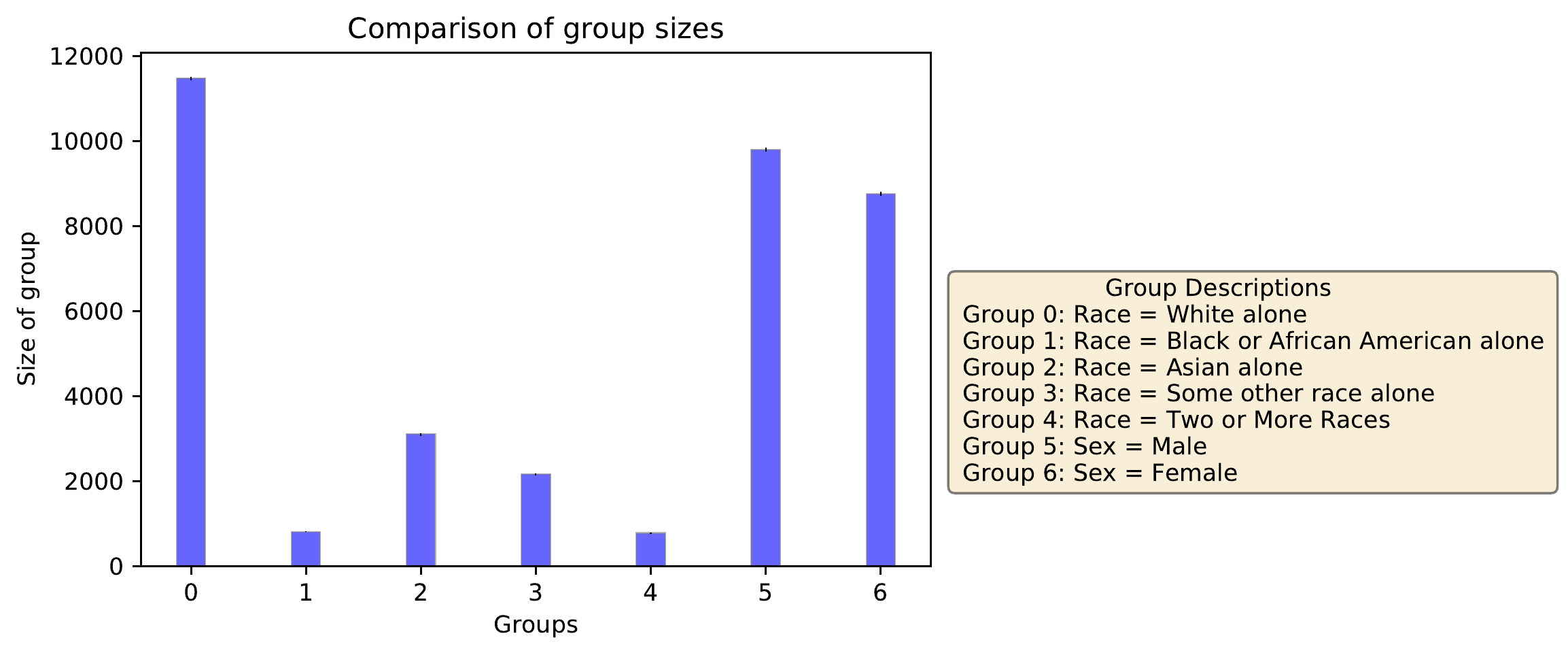}
\caption{The first plot shows the median over 100 independent trials of the marginal coverage conditional on membership in each group. The second plot shows the median of average interval width conditional on membership in each group. The third plot shows the average group size (number of elements in each group) over all 100 trials. Details about groups are to the right of each plot. 
The error bars represent 25th and 75th quantiles, and they are not easily visible in the first and third plot as they are quite narrow.}
\label{fig:groupwise-folktables}
\end{figure}

\paragraph{Results} We run 100 independent trials of our experiment with random subsampling of training and test data from the Folktables dataset. The results are shown in Figure \ref{fig:groupwise-folktables}. While MVP obtains the desired coverage across all groups, the naive split conformal prediction method under-covers on some groups and over-covers on others, and the method of \cite{barber2019limits} significantly over-covers on some groups ($G_3, G_4$ and $G_6$). Additionally, MVP consistently predicts smaller-width prediction intervals in comparison to both other methods.


\subsection{Covariate Shift}
\label{sec:exp-shift}


\subsubsection{Known Covariate Shift with UCI Airfoil Data}
We first study the setting of known covariate shift considered by \cite{tibshirani2019conformal} (which introduced the weighted split conformal prediction method that we use as our point of comparison) and replicate their design. Following \cite{tibshirani2019conformal}, we use the airfoil dataset from the UCI Machine Learning Repository \citep{Dua:2019} which consists of data of NACA 0012 airfoils. The dataset contains $N = 1503$ total instances of $d = 5$ features (frequency in Hz, angle of attack in degrees, chord length in meters, free-stream velocity in meters per second, and suction side displacement thickness in meters)
 The target feature for prediction is scaled sound pressure in decibels. In this setting, the data available for calibration is drawn from a different distribution from the data that is used for evaluation, but the distributions differ only in their relative weighting of feature vectors, and the relative weightings (likelihood ratios) are known. 

Weighted split conformal prediction uses the likelihood ratios between the training and evaluation distributions to find weighted quantiles of the conformal scores on the evaluation data distribution. We note that MVP can also make use of these likelihood ratios when they are known. We do so by ``warm starting'' MVP by running it on the data that weighted split conformal prediction uses for calibration, but re-sampled with replacement using rejection sampling according to the known likelihood ratios\footnote{We could have similarly reweighted the data in our potential function using the likelihood ratios, but we choose this method instead so as to apply our algorithm as a black box.}. 


Following the protocol in \cite{tibshirani2019conformal}, for both methods, we use 25\% of the data to train the underlying linear regression model that will be given to both MVP and weighted split conformal prediction. (It is necessary to use a separate split of the data for the method of \cite{tibshirani2019conformal}, but for our method we could have shared data between training and calibration, which would give us an advantage of the sort we demonstrated in Section \ref{sec:exp-exchange}. We do not do this in this experiment to disentangle different aspects of the comparison between our techniques). The weighted split conformal prediction algorithm is then given a calibration dataset of 25\% of the data to compute the residual quantiles and finally samples with replacement 50\% of the remaining points for the evaluation dataset, with probabilities proportional to $w(x) = \exp(x^T \beta)$, where $\beta = (-1, 0, 0, 0, 1)$. This final fold simulates a synthetic covariate shift and in our comparison the  weighted split conformal method that has oracle access to the shift likelihood ratios. When running MVP, we use this 25\% of the dataset in a comparable way: we sample the calibration fold of the remaining data with replacement with probilities proportional to $w(x)$ and use it to run MVP as a warm start (i.e. the predictions that MVP makes on this fold are not recorded in the metrics we report). This uses the known conformal scores in a similar way to how they are used in weighted split conformal prediction.  MVP is then evaluated on an evaluation dataset obtained the same way as for weighted split conformal, by sampling 50\% of the remaining data with probabilities proportional to $w(x)$. We run MVP with $m=40$ threshold-calibration buckets. 

\begin{figure}
\includegraphics[width=0.5\linewidth]{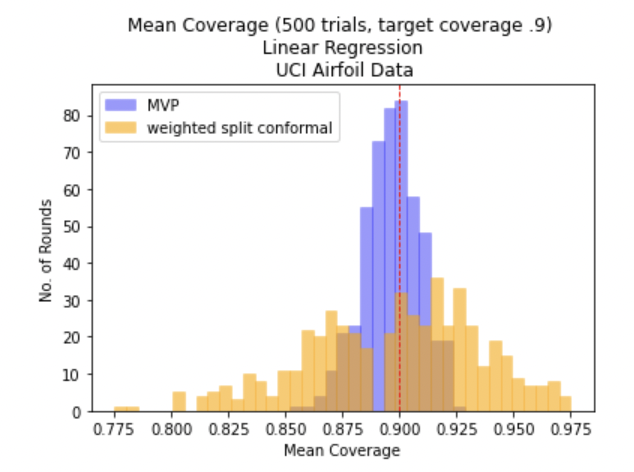} \includegraphics[width=0.5\linewidth]{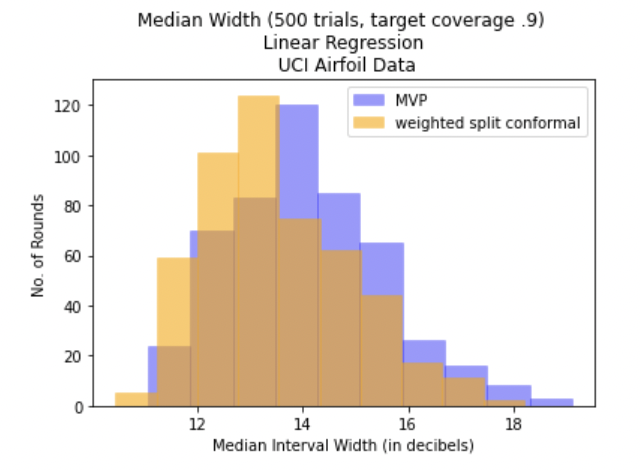}
\caption{The left-hand figure shows a histogram of the coverage rate of MVP and weighted split conformal prediction over 500 trials; the right-hand figure is a histogram of the median prediction interval widths over the same 500 trials.}
\label{fig:shift1}
\end{figure}

Figure \ref{fig:shift1} shows a histogram of the coverage rate and median prediction interval width of both methods  over 500 trials of the experiment, where each trial indicates a different train-test split of the data and a different sampling of the shifted data for the algorithm. We see that MVP obtains coverage that is significantly more tightly concentrated around its target (0.9) compared  to weighted split conformal prediction and comparable interval widths. We note that this is even without letting MVP train its regression model on the calibration dataset. 


\subsubsection{Unknown Shift with Folktables Data}
Next we evaluate split conformal prediction and MVP on real data exhibiting distribution shift, in which the distribution changepoint and propensity scores are unknown (and so cannot be used to weight the calibration set as in our earlier experiment). Here we use the Folktables package  \citep{ding2021retiring}.  The dataset consists of $N = 263973 \text{ (CA: }195665, \text{PA: }68308)$  instances each comprising of $d = 9$ features. We subsample the dataset ($.4$ of CA, $.2$ of PA), thus using $N = 91927$ overall. The features of the data are Census demographic attributes and the target prediction variable is income. We follow \cite{ding2021retiring} in investigating covariate shift that results from using data derived from different states related to the same task.

\begin{figure}
\includegraphics[width=0.5\linewidth]{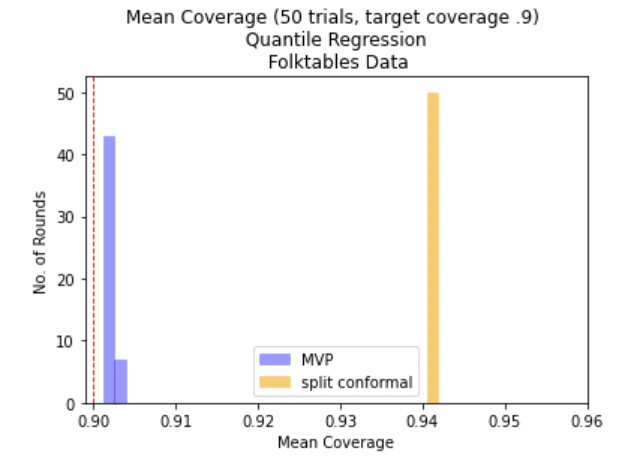} 
\includegraphics[width=0.5\linewidth]{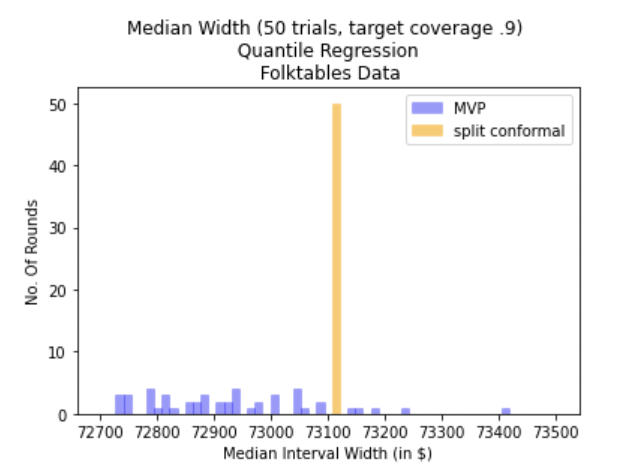}
\caption{The left-hand figure shows a histogram of the coverage for MVP and split conformal prediction over 50 trials; the right-hand figure shows a histogram of the prediction interval width. }
\label{fig:shift2}
\end{figure}

As in all of our experiments, MVP is trained using $m=40$ buckets for calibration. For both MVP and split conformal prediction we use the quantile-regression based conformal score from \cite{romano2019conformalized}, using a quantile regression model trained on half of the available California data. 
We then use the remaining California data as the calibration dataset, used to ``warm start" MVP and compute the residual quantiles for split conformal prediction. Finally, we evaluate MVP and split conformal on the Pennsylvania data and report a histogram of the empirical coverage and interval widths for both methods over 50 trials in Figure \ref{fig:shift2}. MVP comes very close to its coverage target (0.9), whereas split conformal prediction significantly over-covers. Similarly, MVP obtains narrower average prediction interval widths. Here the empirical coverage for both methods is much more tightly concentrated than it is for the UCI Airfoil dataset: this is because the dataset we are using in this experiment is roughly 60 times larger. 

\subsection{Time Series Data}
\label{sec:exp-time}
\subsubsection{Basic Experimental Setup and Marginal Coverage}

In this set of experiments, we run MVP on stock market data and compare our performance to  the Adaptive Conformal Inference (ACI) algorithm of  \cite{gibbs2021adaptive}, a recent method that guarantees \emph{marginal} coverage for adversarially chosen data. In contrast to MVP, ACI promises only marginal coverage (in particular, its guarantees are not threshold calibrated), and so we expect its convergence to be faster but that its thresholds will fluctuate more; our experiments bear this out. 

\begin{figure}
\centering
\includegraphics[width=0.8\linewidth]{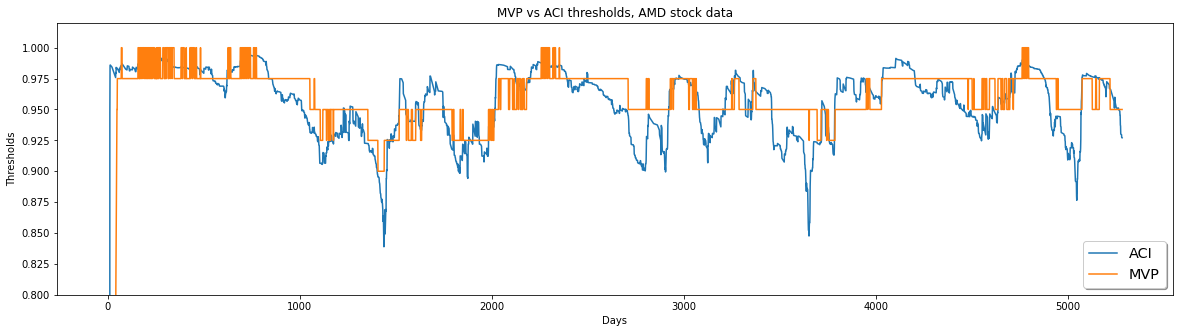}
\caption{A single trajectory of ACI and MVP thresholds plotted together; for convenience, threshold values are only displayed once MVP and ACI thresholds have risen above 0.8. One can see that MVP and ACI trajectories have somewhat similar shapes, but MVP exhibits a more stable behavior.}
\label{fig:singlegroup_MVP_ACI}
\end{figure}

To directly compare to ACI, we use the same dataset and model construction as in \cite{gibbs2021adaptive}. Specifically, we start with WSJ daily open price data\footnote{Available at \url{https://www.wsj.com/market-data}} for AMD stock in 2000-2020 (corresponding to $T = 5283$ price points $p_1, \ldots, p_T$). We calculate daily returns $r_t$ as $r_t = \frac{p_t-p_{t-1}}{p_{t-1}}$ for every day $t$. Based on the returns, we then calculate the (realized) daily volatility as $v_t = r^2_t$ for $t \in [T]$. For our  prediction task we train a model to estimate daily volatility levels $v_t$. Following \cite{gibbs2021adaptive}, we use a standard sequential prediction model called GARCH~\citep{garch_paper}; every day, GARCH makes volatility prediction $\sigma_t$, and autoregressively updates the model once it sees the realized volatility $v_t$. The conformal score we use on day $t$ is the \emph{normalized regression score} $s_t(t,v) = |v_t - \sigma_t|/\sigma_t$. Here $\sigma_t$ is the prediction that the GARCH model makes at round $t$, and possible realizations of the volatility $v_t$ play the role of the label. We run MVP and ACI, for miscoverage target $\delta = 0.1$, on the (rescaled) scores\footnote{MVP assumes that the input scores $s_t \in [0, 1]$, but in this set of experiments we can only guarantee that the normalized regression score $s_t = |v_t - \sigma_t|/\sigma_t \in [0, \infty)$. Due to this, we feed MVP (and ACI, for consistency) modified scores $\tilde{s}_t  = \frac{s_t}{1+s_t} \in [0, 1]$. This type of rescaling works more generally in any setting where MVP's input scores belong to $[0, \infty)$ and need to be rescaled to be in $[0, 1]$: indeed, observe that the mapping $x \rightarrowtail \frac{x}{1+x}$ is a monotonic continuous bijection from $[0, \infty)$ to $[0, 1)$.} $\tilde{s}_1, \ldots, \tilde{s}_T$ of the GARCH model trained to predict AMD stock volatility. In all our experiments with ACI, we set the ACI hyperparameters as follows: $\gamma = 0.005$ (step size), lookback = 100, offset = 10. Figure~\ref{fig:singlegroup_MVP_ACI} shows the sequences of conformity thresholds for MVP and ACI. In general we find that even when we only measure marginal prediction, MVP performs comparably to ACI. Both methods obtain coverage close to the target rate of $0.9$, where ACI consistently gets a bit closer to the target rate. 
We visually observe that MVP makes more stable predictions compared to ACI, locally converging to a small stable set of threshold values (and moving over to the next stable set of thresholds once the scores have drifted sufficiently far), whereas ACI uses continuously fluctuating threshold values (this is expected, since it is not aiming for threshold calibrated coverage). 

\subsubsection{Multigroup Coverage}
Next, we augment the experimental setup to investigate multigroup coverage. We define a set of groups based on whether the index of the trading day is divisible by $1, \ldots, 20$: Define $x_t = t$, and let $\cG = \{G_1, \ldots, G_{20}\}$, where $G_i$ is defined as the set of all $t$ such that $t \equiv 0 \mod i$. In other words, $G_1$ consists of the set of all time steps, $G_2$ consists of even time steps, $G_3$ consists of time-steps that are multiples of 3, and so on. As these sub-groups mutually intersect, it is not possible to run a separate copy of ACI on each one. 

To provide sub-group variability, we artificially introduce varying levels of group-specific additive noise into the stock return data: for each $i \in [20]$, we add noise sampled from $\mathcal{N}(0, \hat{\sigma}_\text{ret})$ to the stock return $r_t$ on all days $t$ that fall into group $G_i$, where $\hat{\sigma}_\text{ret}$ is the empirical standard deviation of the returns sequence. This noise is additive: so the returns on a day that falls into multiple groups are perturbed by the sum of the group-specific perturbations. 

\begin{figure}
\centering
\includegraphics[width=0.8\linewidth]{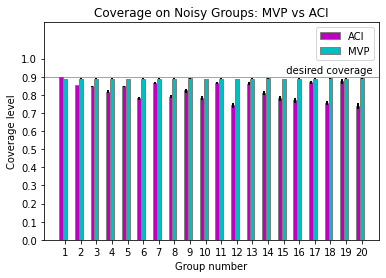}
\caption{MVP and ACI median coverage (over 20 indep.\ trials) on groups 1-20 on noisy data (group $j$ consists of days $t$ such that $t \equiv 0 \mod j$). MVP closely matches desired coverage level on all groups, whereas ACI significantly undercovers (within 10-20\% from the target). In interpreting the plot, note that Group 1 consists of all of the rounds (and so represents overall marginal coverage), and that each group $j$ consists of a $1/j$ fraction of the data, so the groups become increasingly small from left to right. Note the very small (barely visible) error bars (spanning 25th to 75th quantile coverage): For ACI, the largest error quantile width across groups is $0.0303$, whereas for MVP it is even smaller: $0.007$.}
\label{fig:multigroup_MVP_ACI}
\end{figure}

We now run ACI and MVP on the scores produced by GARCH when it is trained on this noisy data. MVP is given the set of groups $\cG$. Figure~\ref{fig:multigroup_MVP_ACI} shows a plot of the median coverage rates (over 20 independent trials) for both ACI and MVP on each of the 20 groups. As expected, MVP achieves close to its target coverage on each group, whereas ACI --- although getting very close to its target marginal coverage (see group 1) undercovers on most other groups, sometimes significantly as a result of the extra added noise.

\subsubsection{Adversarial Ordering} 
\begin{figure}
\begin{subfigure}{0.5\textwidth}
\includegraphics[width=\linewidth]{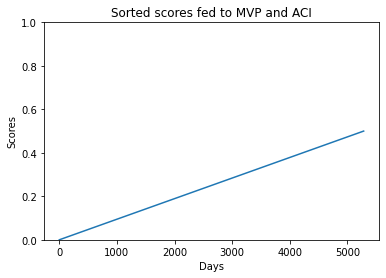}
\caption{}
\label{fig:sorted_scores}
\end{subfigure}
\begin{subfigure}{0.5\textwidth}
\includegraphics[width=\linewidth]{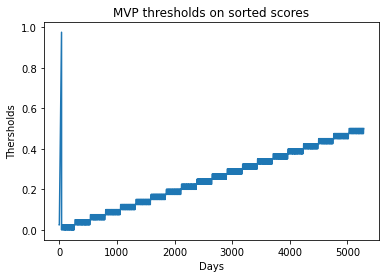}
\caption{}
\label{fig:MVP_thrs_sorted}
\end{subfigure}
\begin{subfigure}{0.5\textwidth}
\includegraphics[width=\linewidth]{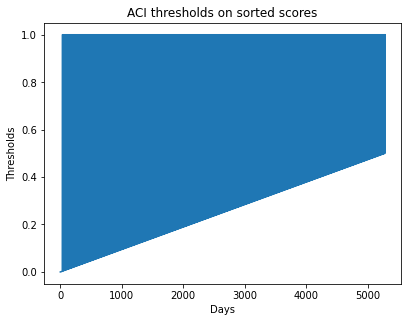}
\caption{}
\label{fig:ACI_actual_thrs_sorted}
\end{subfigure}
\begin{subfigure}{0.5\textwidth}
\includegraphics[width=\linewidth]{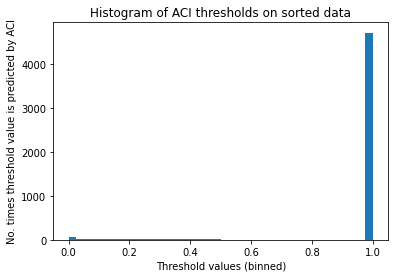}
\caption{}
\label{fig:ACI_thrs_sorted}
\end{subfigure}
\caption{MVP and ACI behavior on a sequence of sorted scores. Figure (a) plots the sequence of scores fed to both MVP and ACI. Figure (b) plots the sequence of thresholds chosen by MVP --- note that it closely tracks the sequence of scores. Figure (c) plots the sequence of thresholds chosen by ACI. It appears to fill the upper diagonal region because it fluctuates so rapidly between the maximum value (1) and just below the score sequence. Figure (d) gives a histogram for the thresholds chosen by ACI, showing that ACI is almost always choosing the uninformative maximum threshold.}
\label{fig:sorted}
\end{figure}

Finally, we present an experiment which tests MVP and ACI on a fully adversarial sequence of conformal scores: a  sequence that linearly grows from $0.0$ to $0.5$ in $T = 5283$ equal steps\footnote{That is, the sequence is $\{\frac{0.5 \cdot i}{T-1}\}_{t=0}^{T-1}$, where $T = 5283$ is chosen to be the same as in the above experiments}, as shown in Figure~\ref{fig:sorted_scores}. In this ordering, the next score is always larger than the algorithm has ever seen before --- hence traditional conformal prediction methods that rely on the exchangeability assumption would obtain 0 coverage on this sequence.

As expected, ACI struggles when it sees scores that are always increasing. 
ACI and MVP are both guaranteed to approach the target marginal coverage, but this sequence serves to elucidate the difference between simple marginal coverage and threshold calibrated coverage. The trajectory of ACI's predicted thresholds on all rounds is shown in Figure~\ref{fig:ACI_actual_thrs_sorted}. ACI's threshold oscillates rapidly between just below the current score and the maximum value (1) that its trajectory appears to fill the space between the score sequence and 1. We also show the histogram of ACI's thresholds over its full trajectory, which shows that most of them in fact correspond to the trivial prediction interval corresponding to the maximum threshold value. This reveals that on an increasing sequence, ACI obtains its target coverage by using a strategy that is very similar to the uninformative ``cheating'' strategy that we outlined in  Footnote~\ref{footnote:allno}: namely, ACI predicts the trivial prediction interval (all of $[0, 1]$) on most rounds, and periodically tries to predict lower threshold values (on which it miscovers and is forced back into predicting the full interval). These prediction intervals are not threshold calibrated. In contrast, MVP's sequence of predicted thresholds have to be threshold-calibrated hence (as shown in Figure~\ref{fig:MVP_thrs_sorted}) they closely track the actual score sequence, resulting in much more informative coverage intervals. 

Beyond recognizing that ACI's thresholds, as opposed to MVP's, are uninformative in this setting, we can also see a concrete drawback of ACI's strategy by looking at its average prediction set width. Namely, suppose that the linearly increasing sequence of scores represent \emph{regression scores} $s_t(y_t, \hat{y}_t) = |y_t - \hat{y}_t|$ in a simple regression problem. Then, each threshold $q_t$ generated by ACI or MVP will produce an interval of width $2q_t$. In this case, the average width attained by MVP will be 0.526, whereas the average width attained by ACI will be 1.839. What is more, note that MVP's thresholds closely track the magnitude of the presented sequence of scores, while ACI's threshold is 1 most of the time no matter what subrange of $[0, 1]$ the observed scores are in. Therefore, if we generate increasing scores from $0$ to $b$ (above, we took $b = 0.5$), where $b$ can be set arbitrarily small, we will get examples of adversarial data on which the prediction interval widths of ACI are \emph{arbitrarily worse} than the prediction widths produced by MVP.

\subsection{A Classification Task: ImageNet}
\label{sec:imagenet}
\begin{figure}
\centering
\includegraphics[width=0.7\linewidth]{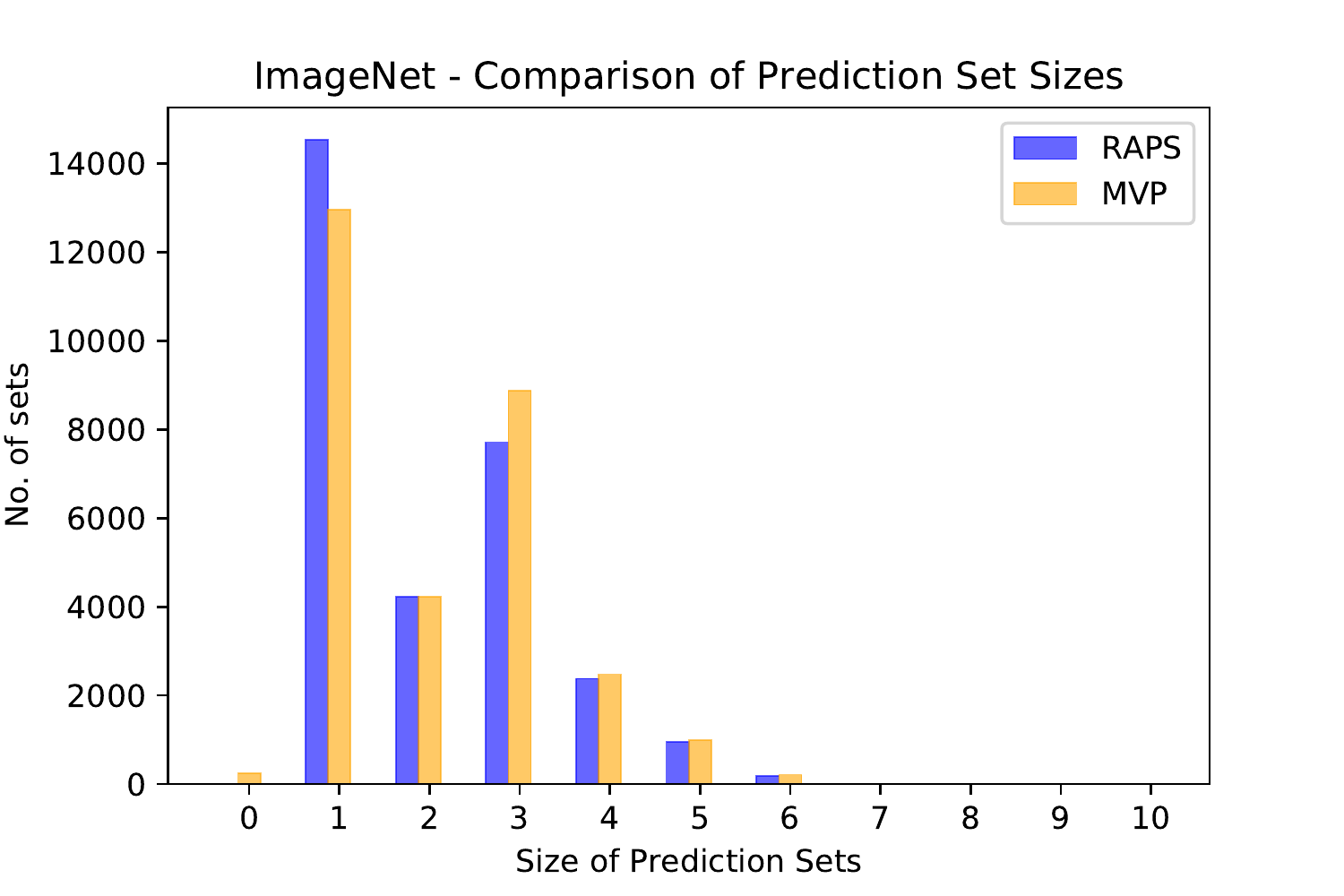}
\caption{A bar graph showing the size of prediction-sets generated by MVP and RAPS over a dataset of 30,000 images. MVP achieves prediction-set sizes on par with RAPS.}
\label{fig:imgnet-pred-set-size}
\end{figure}

In this section, we compare the performance of MVP against an existing conformal prediction method for the task of generating prediction sets in image classification. The recent work of \cite{angelopoulos2020uncertainty} details and implements an algorithm, \textit{Regularized Adaptive Prediction Sets} (RAPS) which, given a trained image classifier, generates small-sized prediction sets of image labels with marginal coverage guarantees. This is done by defining a modified conformal score which empirically produces smaller and more stable sets compared to previously used scores \citep{romano2020classification}. \\ \\
Using ResNet-152 as the base image classifier, we use calibration data of size 1000 from ImageNet to train RAPS. This same data is used as a "warm-start" training set for MVP (i.e. MVP predicts sets for this data and uses it to update variables used in the algorithm; MVP's performance over these time-steps is not recorded). MVP is run with $m = 40$ calibration buckets.  The results shown in Figure~\ref{fig:imgnet-pred-set-size} detail the performance of both methods (using the same conformal score) on a held-out validation dataset of size 30,000.

\paragraph{Results} The marginal coverage achieved by RAPS across all $T = 30000$ images is 0.90523, and the marginal coverage achieved by MVP is 0.902133. The average prediction-set size for RAPS and MVP are 2.0506 and 2.13986 respectively, and the distribution across prediction-set sizes is similar for both methods. Once again we achieve competitive performance with ``traditional'' state of the art conformal prediction methods, even in a setting favorable to them (i.e. a setting with i.i.d. data in which only marginal coverage is measured).

\paragraph{Acknowledgements}
This research was supported in part by the Simons Foundation Collaboration on Algorithmic Fairness and NSF grant FAI-2147212. We thank Stephen Bates for helpful comments on an early version of this paper. 

\bibliographystyle{plainnat}
\bibliography{refs}

\appendix
\section{Omitted Proofs from Section \ref{sec:anal}}
\label{app:anal}
\lemsurrogatelossincrease*
\begin{proof}
Fix $\pi_t = (x_t, s_t, q_t)$. For simplicity, we write $L_t = L_t(\pi_{1:t-1} \oplus \pi_t)$ and $L^{G,i}_t=L^{G,i}_t(\pi_{1:t-1} \oplus \pi_t)$ in the remainder of this proof. For any $(G,i) \not\in A_t(\pi_t)$, we have that 
\begin{align*}
    V_{t}^{G,i} &= V_{t-1}^{G,i}\\
    n_{t}^{G,i} &= n_{t-1}^{G,i}.
\end{align*}
and hence $L^{G,i}_t = L^{G,i}_{t-1}$.

On the other hand, for any $(G,i) \in A_t(\pi_t)$, we have
\begin{align*}
    V_{t}^{G,i} &= V_{t-1}^{G,i} + v_{\delta}(q_{t}, s_{t})\\
    n_{t}^{G,i} &= n_{t-1}^{G,i} + 1.
\end{align*}

Then, we can bound the change in loss for that group-bucket pair $(G,i) \in A_t(\pi_t)$ in the following way:
\begin{align*}
    &L_{t}^{G,i} - L_{t-1}^{G,i} \\
    &= \left(\exp\left(\eta \frac{V_{t}^{G,i}}{f(n_{t}^{G,i})}\right) + \exp\left(-\eta \frac{V_{t}^{G,i}}{f(n_{t}^{G,i})}\right) \right) - \left(\exp\left(\eta \frac{V_{t-1}^{G,i}}{f(n_{t-1}^{G,i})}\right) + \exp\left(-\eta \frac{V_{t-1}^{G,i}}{f(n_{t-1}^{G,i})}\right) \right) \\
    &= \left(\exp\left(\eta \frac{V_{t-1}^{G,i}  +v_{\delta}(q_{t}, s_{t})}{f(n_{t-1}^{G,i}+1)}\right) + \exp\left(-\eta \frac{V_{t-1}^{G,i} + v_{\delta}(q_{t}, s_{t})}{f(n_{t-1}^{G,i}+1)}\right) \right) - \\ & ~~~~ \left(\exp\left(\eta \frac{V_{t-1}^{G,i}}{f(n_{t-1}^{G,i})}\right) + \exp\left(-\eta \frac{V_{t-1}^{G,i}}{f(n_{t-1}^{G,i})}\right) \right) \\
    &\numrel{\leq}{ineq:monotone} \left(\exp\left(\eta \frac{V_{t-1}^{G,i}  +v_{\delta}(q_{t}, s_{t})}{f(n_{t-1}^{G,i})}\right) + \exp\left(-\eta \frac{V_{t-1}^{G,i} + v_{\delta}(q_{t}, s_{t})}{f(n_{t-1}^{G,i})}\right) \right) - \\ & ~~~~ \left(\exp\left(\eta \frac{V_{t-1}^{G,i}}{f(n_{t-1}^{G,i})}\right) + \exp\left(-\eta \frac{V_{t-1}^{G,i}}{f(n_{t-1}^{G,i})}\right) \right) \\
    &= \exp\left(\eta \frac{V_{t-1}^{G,i}}{f(n_{t-1}^{G,i})}\right)\left(\exp\left(\eta \cdot \frac{v_{\delta}(q_{t}, s_{t})}{f(n_{t-1}^{G,i})}\right) - 1\right) + \exp\left(-\eta \frac{V_{t-1}^{G,i}}{f(n_{t-1}^{G,i})}\right)\left(\exp\left(-\eta\cdot \frac{v_{\delta}(q_{t}, s_{t})}{f(n_{t-1}^{G,i})}\right) - 1\right) \\
    &\numrel{\leq}{ineq:taylor} \exp\left(\eta \frac{V_{t-1}^{G,i}}{f(n_{t-1}^{G,i})}\right)\left(\eta \cdot \frac{v_{\delta}(q_{t}, s_{t})}{f(n_{t-1}^{G,i})} + \frac{2\eta^2}{f(n_{t-1}^{G,i})^2}\right) +\exp\left(-\eta \frac{V_{t-1}^{G,i}}{f(n_{t-1}^{G,i})}\right)\left(-\eta \cdot \frac{v_{\delta}(q_{t}, s_{t})}{f(n_{t-1}^{G,i})} +  \frac{2\eta^2}{f(n_{t-1}^{G,i})^2}\right) \\
    &= \eta \cdot \frac{v_{\delta}(q_{t}, s_{t})}{f(n_{t-1}^{G,i})} \left(\exp\left(\eta \frac{V_{t-1}^{G,i}}{f(n_{t-1}^{G,i})}\right) - \exp\left(-\eta \frac{V_{t-1}^{G,i}}{f(n_{t-1}^{G,i})}\right)\right) + \\ & ~~~~  \frac{2\eta^2}{f(n_{t-1}^{G,i})^2} \left(\exp\left(\eta \frac{V_{t-1}^{G,i}}{f(n_{t-1}^{G,i})}\right) +  \exp\left(-\eta \frac{V_{t-1}^{G,i}}{f(n_{t-1}^{G,i})}\right)\right) \\
    &= \eta v_{\delta}(q_{t}, s_{t}) C_{t-1}^{G,i} + \frac{2\eta^2}{f(n_{t-1}^{G,i})^2} L_{t-1}^{G, i}
\end{align*}

The first inequality \eqref{ineq:monotone} holds due to $e^x + e^{-x}$ being monotone with respect to $|x|$,  inequality \eqref{ineq:taylor} follows from the fact that for $0 \leq |x| \leq \frac{1}{2}$, $\exp(x) \leq 1 + x + 2x^2$, and $\left|\eta\cdot \frac{v_{\delta}(q_{t}, s_{t})}{f(n_{t-1}^{G,i})}\right| \le \frac{1}{2}$ because of the way we set $\eta \in (0,1/2)$ and the fact that $|v_{\delta}(q_t, s_t)| \in [0,1]$ and $f(n^{G,i}_{t-1}) \ge 1$. 

Therefore, we have
\begin{align*}
    L_t - L_{t-1} &= \sum_{(G,i) \in A_t(\pi_t)} L^{G,i}_t - L^{G,i}_{t-1} \\
    &\le \sum_{(G,i) \in A_t(\pi_t)} \eta v_{\delta}(q_{t+1}, s_{t+1}) C_{t-1}^{G,i} + \frac{2\eta^2}{f(n_t^{G,i})^2} L_{t}^{G, i} 
\end{align*}
\end{proof}

\lemalgboundedincreasesurrogateloss*
\begin{proof}
 For simplicity, suppose we write \[ u(q,s) =  v_\delta(q, s) \sum_{(G,i) \in A_t(\pi_t)} C^{G,i}_t = v_\delta(q, s) C^{q}_t
\]
where we overload the notation to write
\[
    C^{q}_t = C^{\Binv(q)}_t = \sum_{G \in \cG(x_t)} C^{G,\Binv(q)}_t.
\]

\paragraph{Case (i) $C^{i}_{t-1} < 0$ for all $i \in [n]$:}
With $q_t=1$, we have
\begin{align*}
    \E_{q_{t} \sim Q^L, s_{t} \sim Q^A}[u(q_t, s_t) | x_t] = C^{1}_{t-1}(x_{t}) \E_{s_t \sim Q^A}[v_\delta(1, s_t) | x_t] < 0
\end{align*}
as $v_{\delta}(1, d_t) = 1 - (1-\delta) > 0$.

\paragraph{Case (ii) $C^{i}_{t-1} > 0$ for all $i \in [n]$:}
With $q_t=0$, we have
\begin{align*}
    \E_{q_{t} \sim Q^L, s_{t} \sim Q^A}[u(q_t, s_t) |x_t] = C^{0}_{t-1}(x_{t}) \E_{s_t \sim Q^A}[v_\delta(0, s_t) |x_t] < \rho C^{0}_{t-1}(x_{t}) < \rho L_{t-1}.
\end{align*}
as we have
\begin{align*}
    \E_{s_t \sim Q^A}[\cover(0, s_t)|x_t] - (1-\delta)
    \le \E_{s_t \sim Q^A}[\cover(0, s_t)|x_t] = \Pr_{s_t \sim Q^A}[s_t = 0|x_t] \le \rho
\end{align*}
\paragraph{Case (iii) there exists $i^* \in [n-1]$ such that $C^{i^*}_{t-1} \cdot C^{i^*+1}_{t-1} \le 0$:}

First, consider the case where $C^{i^*}_{t-1} \ge 0$ and $C^{i^*+1}_{t-1} \le 0$. Then, we have
\begin{align*}
    &\E_{q_{t} \sim Q^L, s_{t} \sim Q^A}[u(q_t, s_t) |x_t]\\
    &= p_t \E_{s_t \sim Q^A}\left[u\left(\frac{i^*}{n} - \frac{1}{rn}, s_t\right) \middle| x_t \right] + (1-p_t) \E_{s_t \sim Q^A}\left[u\left(\frac{i^*}{n}, s_t\right) \middle| x_t\right]\\
    &= p_t C^{i^*}_{t-1}(x_{t}) \E_{s_t \sim Q^A}\left[ v_\delta\left(\frac{i^*}{n} - \frac{1}{rn}, s_t\right) \middle| x_t \right] + (1-p_t) C^{i^*+1}_{t-1} \E_{s_t \sim Q^A}\left[ v_\delta\left(\frac{i^*}{n}, s_t\right) \middle| x_t \right]\\
    &\le p_t C^{i^*}_{t-1} \left(\E_{s_t \sim Q^A}\left[ v_\delta\left(\frac{i^*}{n}, s_t\right) \middle| x_t \right] \right) + (1-p_t) C^{i^*+1}_{t-1} \E_{s_t \sim Q^A}\left[ v_\delta\left(\frac{i^*}{n}, s_t\right) \middle| x_t \right]\\
    &= \E_{s_t \sim Q^A}\left[ v_\delta\left(\frac{i^*}{n}, s_t\right) \middle| x_t \right]  \left(p_t C^{i^*}_{t-1}   + (1-p_t) C^{i^*+1}_{t-1} \right)\\
    &= 0.
\end{align*}
The first inequality follows from the fact that $\cover(\frac{i^*}{n} - \frac{1}{rn}, s) \le \cover(\frac{i^*}{n}, s)$ for any $s$.
Now, consider the other case where $C^{i^*}_{t-1} \le 0$ and $C^{i^*+1}_{t-1} \ge 0$.
\begin{align*}
    &\E_{q_{t} \sim Q^L, s_{t} \sim Q^A}\left[u(q_t, s_t) \middle| x_t \right]\\
    &= p_t \E_{s_t \sim Q^A}\left[u\left(\frac{i^*}{n} - \frac{1}{rn}, s_t\right) \middle| x_t \right] + (1-p_t) \E_{s_t \sim Q^A}\left[u\left(\frac{i^*}{n}, s_t\right) \middle| x_t \right]\\
    &= p_t C^{i^*}_{t-1}(x_{t}) \E_{s_t \sim Q^A}\left[ v_\delta\left(\frac{i^*}{n} - \frac{1}{rn}, s_t\right) \middle| x_t \right] + (1-p_t) C^{i^*+1}_{t-1} \E_{s_t \sim Q^A}\left[ v_\delta\left(\frac{i^*}{n}, s_t\right) \middle| x_t \right]\\
    &\le p_t C^{i^*}_{t-1}(x_{t}) \E_{s_t \sim Q^A}\left[ v_\delta\left(\frac{i^*}{n} - \frac{1}{rn}, s_t\right)\middle| x_t \right] + (1-p_t) C^{i^*+1}_{t-1} \left(\E_{s_t \sim Q^A}\left[ v_\delta\left(\frac{i^*}{n} - \frac{1}{rn}, s_t\right)\middle| x_t \right] + \rho \right)\\
    &= \rho (1-p_t) C^{i^*+1}_{t-1} +  \E_{s_t \sim Q^A}\left[ v_\delta\left(\frac{i^*}{n} - \frac{1}{rn}, s_t\right)\middle| x_t\right]  \left(p_t C^{i^*}_{t-1}   + (1-p_t) C^{i^*+1}_{t-1} \right)\\
    &= \rho L_{t-1}.
\end{align*}
The first inequality follows from the fact that
\begin{align*}
&\Pr_{s_t \sim Q^A}\left[\cover\left(\frac{i^*}{n}, s_t\right)\middle| x_t\right] - \Pr_{s_t \sim Q^A}\left[\cover\left(\frac{i^*}{n} - \frac{1}{rn}, s_t\right)\middle| x_t\right]  \\
&\le \Pr_{s_t \sim Q^A}\left[s_t \in \left[\frac{i^*}{n} - \frac{1}{rn}, \frac{i^*}{n}\right]\middle| x_t \right] \\
&\le \rho.
\end{align*}

\end{proof}

\thmalgmutivalidguarantee*
\begin{proof}
Fix any round $t \in [T]$ and transcript $\pi_{1:t-1}$. For simplicity, we write $L_t = L_t(\pi_{1:t})$. Then, we can
use Lemma~\ref{lem:surrogate-loss-increase}  to prove the following lemma.

\begin{lemma}
\label{lem:multiplicative-increase}
Fix any transcript $\pi_{1:T}$. Then, for any round $t \in [T]$, we have
\begin{align*}
    L_{t} \le L_{t-1}\left(1+\frac{\eta v_{\delta}(q_{t}, (x_t, s_t))}{L_{t-1}}\sum_{(G,i) \in A_t(\pi_t)} C_{t-1}^{G,i} + \sum_{(G,i) \in A_t(\pi_t)} \frac{2\eta^2}{f(n_t^{G,i})^2}\right).
\end{align*}
\end{lemma}
\begin{proof}
Fix transcript $\pi_{1:T}$. Then at any round $t$, we have 
\begin{align*}
&L_t \\
&= L_{t-1} + L_t - L_{t-1}  \\
&\leq L_{t-1} +  \sum_{(G,i) \in A_t(\pi_t)} \eta v_{\delta}(q_{t}, (x_t, s_t)) C_{t-1}^{G,i} + \frac{2\eta^2}{f(n_t^{G,i})^2} L_{t-1}^{G, i}  \quad&&\text{(Lemma \ref{lem:surrogate-loss-increase})}\\
&\leq L_{t-1} + \sum_{(G,i) \in A_t(\pi_t)} \eta v_{\delta}(q_{t}, (x_t, s_t)) C_{t-1}^{G,i} + L_{t-1}\sum_{(G,i) \in A_t(\pi_t)} \frac{2\eta^2}{f(n_t^{G,i})^2} \quad&&\left( L^{G,i}_{t-1} \le L_{t-1}\right)\\
&\leq L_{t-1}\left(1 +\frac{\eta v_{\delta}(q_{t}, (x_t, s_t))}{L_{t-1}}\sum_{(G,i) \in A_t(\pi_t)} C_{t-1}^{G,i} +  \sum_{(G,i) \in A_t(\pi_t)}\frac{2\eta^2}{f(n_t^{G,i})^2} \right). 
\end{align*}
\end{proof}

Applying Lemma~\ref{lem:multiplicative-increase} recursively, we get
\begin{align*}
L_T &\le L_0 \prod_{t=1}^T \left(1+\frac{\eta v_{\delta}(q_{t}, (x_t, s_t))}{L_{t-1}}\sum_{(G,i) \in A_t(\pi_t)} C_{t-1}^{G,i} + \sum_{(G,i) \in A_t(\pi_t)} \frac{2\eta^2}{f(n_t^{G,i})^2} \right)\\
&\numrel{\le}{ineq:exp} L_0 \prod_{t=1}^T \exp\left(\frac{\eta v_{\delta}(q_{t}, (x_t, s_t))}{L_{t-1}}\sum_{(G,i) \in A_t(\pi_t)} C_{t-1}^{G,i} + \sum_{(G,i) \in A_t(\pi_t)} \frac{2\eta^2}{f(n_t^{G,i})^2} \right)\\
&\le L_0 \exp\left(\sum_{t=1}^T \frac{\eta v_{\delta}(q_{t}, (x_t, s_t))}{L_{t-1}}\sum_{(G,i) \in A_t(\pi_t)} C_{t-1}^{G,i} + \sum_{t=1}^T\sum_{(G,i) \in A_t(\pi_t)} \frac{2\eta^2}{f(n_t^{G,i})^2} \right)\\
&\le L_0 \exp\left(\sum_{t=1}^T \frac{\eta v_{\delta}(q_{t}, (x_t, s_t))}{L_{t-1}}\sum_{(G,i) \in A_t(\pi_t)} C_{t-1}^{G,i} + \sum_{G \in \cG, i \in [m]} \sum_{n=1}^{n^{G,i}_T} \frac{2\eta^2}{f(n)^2} \right)\\
&\le L_0 \exp\left(\sum_{t=1}^T \frac{\eta v_{\delta}(q_{t}, (x_t, s_t))}{L_{t-1}}\sum_{(G,i) \in A_t(\pi_t)} C_{t-1}^{G,i} + \sum_{G \in \cG, i \in [m]} \sum_{n=1}^{\infty} \frac{2\eta^2}{f(n)^2} \right)\\
&\le L_0 \exp\left(\sum_{t=1}^T \frac{\eta v_{\delta}(q_{t}, (x_t, s_t))}{L_{t-1}}\sum_{(G,i) \in A_t(\pi_t)} C_{t-1}^{G,i} + 2\eta^2K_{\epsilon}|\cG|m\right)\\
&= 2|\cG|m \exp\left(\sum_{t=1}^T \frac{\eta v_{\delta}(q_{t}, (x_t, s_t))}{L_{t-1}}\sum_{(G,i) \in A_t(\pi_t)} C_{t-1}^{G,i} + 2\eta^2K_{\epsilon}|\cG|m\right)
\end{align*}
where inequality \eqref{ineq:exp} follows from $1+x \le \exp(x)$.

Taking the log of both sides, we have 
\begin{align*}
    \ln(L_T) \le \ln(2|\cG|m) + \sum_{t=1}^T \frac{\eta v_{\delta}(q_{t}, (x_t, s_t))}{L_{t-1}}\sum_{(G,i) \in A_t(\pi_t)} C_{t-1}^{G,i} + 2\eta^2K_{\epsilon}|\cG|m
\end{align*}
for any $\pi_{1:T}$.

By Observation \ref{obs:multicoverage}, it suffices to upper bound $\max_{G \in \cG, i \in [m]}\frac{|V^{G,i}_T|}{f(n^{G,i}_T)}$. We have:
\begin{align*}
\max_{G \in \cG, i \in [m]}\frac{|V^{G,i}_T|}{f(n^{G,i}_T)} &= \frac{1}{\eta}\ln\left(\exp\left(\max_{G \in \cG, i \in [m]}\frac{\eta|V^{G,i}_T|}{f(n^{G,i}_T)}\right)\right)\\
&= \frac{1}{\eta}\ln\left(\max_{G \in \cG, i \in [m]} \exp\left(\frac{\eta|V^{G,i}_T|}{f(n^{G,i}_T)}\right)\right)\\
&\le \frac{1}{\eta}\ln\left(\sum_{G \in \cG, i \in [m]} \exp\left(\frac{\eta|V^{G,i}_T|}{f(n^{G,i}_T)}\right)\right)\\
&\le \frac{1}{\eta}\ln\left(\sum_{G \in \cG, i \in [m]} \exp\left(\frac{\eta V^{G,i}_T}{f(n^{G,i}_T)}\right) + \exp\left(\frac{-\eta V^{G,i}_T}{f(n^{G,i}_T)}\right)\right)\\
&= \frac{\ln(L_T)}{\eta}\\
&\le \frac{1}{\eta}\left(\ln(2|\cG|m) + \sum_{t=1}^T \frac{\eta v_{\delta}(q_{t}, (x_t, s_t))}{L_{t-1}}\sum_{(G,i) \in A_t(\pi_t)} C_{t-1}^{G,i} + 2\eta^2K_{\epsilon}|\cG|m\right).
\end{align*}

Taking expectation over $\pi_{1:T}$ on both sides, we get
\begin{align*}
    &\E_{\pi_{1:T}}\left[\max_{G \in \cG, i \in [m]}\frac{|V^{G,i}_T|}{f(n^{G,i}_T)} \right]\\
    &\le \E_{\pi_{1:T}}\left[\frac{1}{\eta}\left(\ln(2|\cG|m) + \sum_{t=1}^T \frac{\eta v_{\delta}(q_{t}, (x_t, s_t))}{L_{t-1}}\sum_{(G,i) \in A_t(\pi_t)} C_{t-1}^{G,i} + 2\eta^2K_{\epsilon}|\cG|m\right) \right] \\
    &\le \frac{1}{\eta}\left(\ln(2|\cG|m)+ 2\eta^2K_{\epsilon}|\cG|m + \E_{\pi_{1:T}}\left[ \sum_{t=1}^T \frac{\eta v_{\delta}(q_{t}, (x_t, s_t))}{L_{t-1}}\sum_{(G,i) \in A_t(\pi_t)} C_{t-1}^{G,i}\right]\right).
\end{align*}

Let us focus only on the third term:
\begin{align*}
    &\E_{\pi_{1:T}}\left[ \sum_{t=1}^T \frac{\eta v_{\delta}(q_{t}, (x_t, s_t))}{L_{t-1}}\sum_{(G,i) \in A_t(\pi_t)} C_{t-1}^{G,i}\right]\\ &=\E_{\pi_{1:T-1}}\left[\E_{\pi_T}\left[ \sum_{t=1}^T \frac{\eta v_{\delta}(q_{t}, (x_t, s_t))}{L_{t-1}}\sum_{(G,i) \in A_t(\pi_t)} C_{t-1}^{G,i} \middle| \pi_{1:T-1}\right]\right]\\
    &= \E_{\pi_{1:T-1}}\left[\sum_{t=1}^{T-1} \frac{\eta v_{\delta}(q_{t}, (x_t, s_t))}{L_{t-1}}\sum_{(G,i) \in A_t(\pi_t)} C_{t-1}^{G,i} + \frac{\eta}{L_{T-1}}\E_{\pi_T}\left[ v_{\delta}(q_{T}, (x_T, s_T))\sum_{(G,i) \in A_T(\pi_T)} C_{T-1}^{G,i}  \middle| \pi_{1:T-1}\right]\right]\\
    &\numrel{\le}{ineq:alg-lem} \E_{\pi_{1:T-1}}\left[\sum_{t=1}^{T-1} \frac{\eta v_{\delta}(q_{t}, (x_t, s_t))}{L_{t-1}}\sum_{(G,i) \in A_t(\pi_t)} C_{t-1}^{G,i} + \eta\rho \right] \\
    &\le \dots\\
    &\le \eta\rho T
\end{align*}
where inequality \eqref{ineq:alg-lem} comes from Lemma~\ref{lem:alg-bounded-increase-surrogate-loss}.

In other words, we have
\begin{align*}
    \E_{\pi_{1:T}}\left[\max_{G \in \cG, i \in [m]}\frac{|V^{G,i}_T|}{f(n^{G,i}_T)}\right] &\le \frac{1}{\eta}\left(\ln(2|\cG|m)+ 2\eta^2K_{\epsilon}|\cG|m + \eta\rho T\right)\\
    &= \frac{\ln (2|\cG| m)}{\eta} +  2\eta |\cG| m K_\epsilon  + \rho T \\
    &\leq \sqrt{4K_\epsilon |\cG| m \ln(|\cG| m)} + \rho T
\end{align*}
where the last inequality follows from setting $\eta = \sqrt{\frac{\ln(|\cG|m)}{2K_\epsilon |\cG|m}}$.
Note that $\eta < 1/2$ as $2\ln(|\cG|m) < K_\epsilon |\cG|m$ because $K_\epsilon \ge 1$.

\end{proof}

\end{document}